\documentclass[11pt]{article}
\usepackage[margin=0.81in]{geometry}
\usepackage{graphicx,hyperref,comment}
\usepackage{amsmath,amsthm,amsfonts,amssymb,color,hyperref,epstopdf,multirow,enumitem}
\usepackage[usenames,dvipsnames]{xcolor}

\newcommand{\rr}{\mathbb{R}}
\newcommand{\rrplus}{\rr^+}

\newcommand{\ep}{\epsilon}

\newcommand{\ra}{\rightarrow}

\renewcommand{\bold}[1]{\textbf{#1}}
\newcommand{\parabold}[1]{\noindent\bold{#1}}

\newcommand{\comm}[1]{~~~~~~\text{(#1)}}

\DeclareMathOperator*{\argmax}{arg\,max}

\newcommand{\difft}[1]{\frac{\mathsf{d} #1}{\mathsf{d} t}}

\newenvironment{pf}{\begin{proof}[\emph{\textbf{Proof: }}]}{\end{proof}}
\newenvironment{pfof}[1]{\begin{proof}[\emph{\textbf{Proof of #1: }}]}{\end{proof}}
\newtheorem{theorem}{Theorem}
\newtheorem{lemma}[theorem]{Lemma}
\newtheorem{corollary}[theorem]{Corollary}

\newtheorem{example}[theorem]{Example}

\newtheorem{proposition}[theorem]{Proposition}
\newtheorem{definition}[theorem]{Definition}

\newcommand{\calE}{\mathcal{E}}

\newcommand{\calN}{\mathcal{N}}
\newcommand{\calO}{\mathcal{O}}
\newcommand{\calP}{\mathcal{P}}

\newcommand{\calZ}{\mathcal{Z}}

\newcommand{\bba}{\mathbf{a}}
\newcommand{\bbb}{\mathbf{b}}

\newcommand{\bbe}{\mathbf{e}}

\newcommand{\bbp}{\mathbf{p}}
\newcommand{\bbq}{\mathbf{q}}
\newcommand{\bbr}{\mathbf{r}}

\newcommand{\bbu}{\mathbf{u}}

\newcommand{\bbx}{\mathbf{x}}
\newcommand{\bby}{\mathbf{y}}
\newcommand{\bbz}{\mathbf{z}}

\newcommand{\bbA}{\mathbf{A}}

\newcommand{\inner}[2]{\left\langle ~#1 ~,~ #2~\right\rangle}
\newcommand{\trans}{^{\mathsf{T}}}

\newcommand{\uu}{\mathbb{U}}
\newcommand{\yy}{\mathbb{Y}}
\newcommand{\bblt}{\mathbb{L}_2}
\newcommand{\bblte}{\mathbb{L}_{2,e}}
\renewcommand{\inner}[2]{\left\langle #1, #2\right\rangle}
\newcommand{\bbone}{\mathbf{1}}

\newcommand{\hbbp}{\hat{\bbp}}
\newcommand{\hbbx}{\hat{\bbx}}
\newcommand{\hbbq}{\hat{\bbq}}

\title{Online Optimization in Games via Control Theory:\\ Connecting Regret, Passivity and Poincar\'{e} Recurrence}
\author{Yun Kuen Cheung\\Royal Holloway\\University of London
\and
Georgios Piliouras\\Singapore University of\\Technology and Design}
\date{}

\begin{document}

\maketitle

\begin{abstract}
We present a novel control-theoretic understanding of online optimization and learning in games, via the notion of \emph{passivity}.
Passivity is a fundamental concept in control theory, which abstracts energy conservation and dissipation in physical systems.
It has become a standard tool in analysis of general feedback systems, to which game dynamics belong.
Our starting point is to show that all continuous-time Follow-the-Regularized-Leader (FTRL) dynamics,
which include the well-known Replicator Dynamic, are \emph{lossless}, i.e.~it is passive with no energy dissipation.
Interestingly, we prove that passivity implies bounded regret, connecting two fundamental primitives of control theory and online optimization.

The observation of energy conservation in FTRL inspires us to present a family of lossless learning dynamics,
each of which has an underlying energy function with a simple gradient structure. This family is closed under \emph{convex combination};
as an immediate corollary, any convex combination of FTRL dynamics is lossless and thus has bounded regret.
This allows us to extend the framework of Fox and Shamma~\cite{fox2013population} to prove not just global 
asymptotic stability results for game dynamics, but Poincar\'{e} recurrence results as well.
Intuitively, when a lossless game (e.g.~graphical constant-sum game) is coupled with lossless learning dynamics,
their feedback interconnection is also lossless, which results in a pendulum-like energy-preserving recurrent behavior,
generalizing \cite{piliouras2014optimization,GeorgiosSODA18}.
\end{abstract}

\begin{figure*}[htp]
\centering
\begin{tabular}{|c|c|c|}
\hline
& \bold{Lossless Learning Dynamic} & \bold{Physical System}\\
\hline
& Replicator Dynamic,  & \\
\bold{examples} & Follow-the-Regularized-Leader  & gravity \\
& (FTRL) dynamics & \\
\hline
\bold{state} & $\bbq ~=~\int \bbp(t)\,\mathsf{d}t~=~$cumulative payoffs  & $h ~=~ $vertical height\\
\hline
\bold{energy} & $E(\bbq) ~=~$storage function  & $V(h) = $ (negative) potential energy\\
\hline
\bold{gradient of} & $\bbx ~=~ \nabla E(\bbq) ~=~$ & $F ~=~ \nabla V(h) ~=~$\\
\bold{energy} & mixed strategies of agents & gravitational force\\
\hline 
 & convex combination (CC):& linear combination (LC):\\
\bold{invariant} & any CC of storage functions produces & any LC of potential energy function is\\
\bold{property} & a lossless learning dynamic that & a potential energy function that yields \\
&  yields the same CC of mixed strategies & the same LC of gravitational forces\\
\hline 
\bold{another} & \multirow{2}{*}{$\bbp ~=~$instantaneous payoffs} & \multirow{2}{*}{$v ~=~$velocity}\\
\bold{analogue} & &\\
\hline 
\bold{change of} & \multirow{2}{*}{$\int \inner{\bbx}{\bbp}\,\mathsf{d}t$} & \multirow{2}{*}{$\int \inner{F}{v}\,\mathsf{d}t$}\\
\bold{energy value} && \\
\hline
\end{tabular}
\caption{Analogues between lossless online learning dynamic and physical system.
The evolution of a system of learning dynamics can be thought of as capturing the movements of particles, 
thus tools from control theory can find direct application in the study of learning dynamics in games.}\label{fig:physics}
\end{figure*}
\section{Introduction}

Online optimization aims at designing algorithms that can maximize performance in unpredictable and even adversarially evolving environments.
The standard benchmark for success in these environments is minimizing regret,
which is defined as the difference between the accumulated performance of the algorithm and that of the best action in hindsight.
One of the most important achievements of the field has been to establish that such regret minimizing algorithms exist~\cite{Cesa06,Shalev2012}.
Amongst the most well-known such algorithms is the class of Follow-the-Regularized-Leader (FTRL) algorithms,
which include as special cases ubiquitous meta-algorithms such as
Multiplicative Weights Update~\cite{freund1999adaptive,Arora05themultiplicative} and Gradient Descent~\cite{Shalev2012}.
It is well known that such algorithms can achieve $\calO(\sqrt{T})$ regret by employing slowly decreasing step-sizes,
and that this bound is effectively optimal given arbitrary payoff sequences.
When applying such algorithms in games, as the sequence of payoffs becomes more predictable, stronger regret guarantees are possible~\cite{rakhlin2013online,foster2016learning,Syrgkanis:2015:FCR:2969442.2969573,bailey2019fast}.
In continuous-time model, FTRL dynamics are once again optimal achieving bounded regret in general settings~\cite{KM17,GeorgiosSODA18,bailey2019finite}.
Hence both from the perspective of optimization as well as game theory, FTRL dynamics constitute effectively an optimal choice.
 
Control theory, on the other hand, is motivated by a seemingly unrelated set of questions.
It aims to develop methodologies for stabilizing complex processes and machines.
Due to its intrinsic connections to real-world systems, control theory revolves around concepts with a strong grounding in physical systems.
A fundamental property of numerous physical systems is \emph{passivity}, which is typically defined in terms of
energy dissipation, conservation and transformation~\cite{Willem72a,Willem72b,ortega2013passivity}.
Passivity is an ``input-output'' property of a system, and expresses that a system which is supplied with bounded energy can only output bounded energy.
Passive systems come equipped with a \emph{storage function} that accumulates the supplied energy
but perhaps with some loss (cf.~energy loss due to friction in mechanical systems).
Overall, passivity encodes a useful notion of stability,
since such system cannot explode into unpredictable out-of-control motion as it would correspond to unbounded energy output.

Although the fields of online optimization and control theory are both well developed with long and distinct histories, 
their interconnection is still rather nascent.
Online algorithms can be abstractly thought as input-output operators where the input is a stream of payoffs,
and the output is a stream of behavioral outcomes. Both notions of regret and passivity are similar properties of
such input-output algorithms/operators and encode a notion of predictability and stability around a reference frame.
In regret, the reference frame is given by the cumulative payoff of past actions, in passivity by energy level sets.
This raises our first set of questions:
\begin{quote}
\emph{Are there formal connections between regret and passivity?
Moreso, can we interpret the optimal regret of FTRL dynamics from a passivity perspective?
Are there similarly optimal learning dynamics / input-output operators?}
\end{quote}

Any formal connection across the two fields is clearly valuable, as it allows for a fusion of ideas and methodologies between two well-developed fields,
and expedite the progress on areas of interest that are common to both, such as game theory~\cite{Fudenberg98,Cesa06,marden2015game,marden2018game}.
For related issues on the intersection of learning, control theory and games, see \cite{Shamma2020} and the references therein.

\begin{figure}[t]
\centering \includegraphics[scale=0.9]{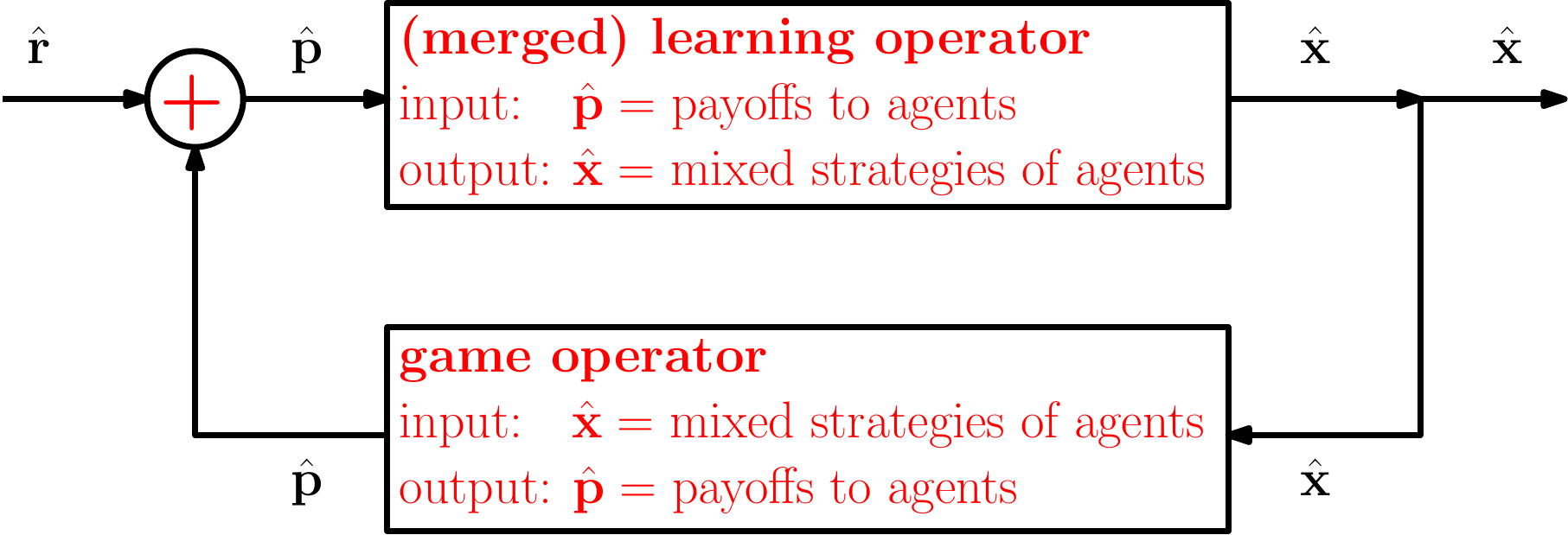}
\caption{A feedback interconnection system that captures a game dynamic, by interconnecting a learning operator and a game operator.
$\hat{\bbr}$ is (random) perturbation to payoffs; in this paper, we only consider the cases where $\hat{\bbr}$ is the zero function.
}\label{fig:FIC-simplified}
\end{figure}

Notably, Fox and Shamma~\cite{fox2013population} proposed a control-theoretic framework for analyzing learning in games.
One of their key contributions is to identify a \emph{modular approach}, where an analysis can be performed by
studying a learning operator (which converts payoff input to strategy output) and a game operator (which converts strategy input to payoff output) \emph{independently},
while the whole game dynamic is a feedback interconnection system of the two operators (see Figure~\ref{fig:FIC-simplified}).
By focusing on coupling learning heuristics that are strictly passive with passive games (e.g.~zero-sum games),
the resulting strictly passive systems were shown to converge to equilibria, generalizing and unifying numerous prior results, e.g.~\cite{hofbauer2009stable}.

The modular approach has allowed numerous works which study learning or game operators separately~\cite{MabrokShamma2016,PSM2018,Mabrok2018,GadjovPavel2019}.
Despite this progress, settings of critical importance for AI such as understanding the perfectly recurrent non-equilibrating behaviors of Gradient Descent (and other FTRL dynamics) 
in zero-sum games has so far remained outside the reach of these techniques~\cite{piliouras2014optimization,GeorgiosSODA18,balduzzi2018mechanics,vlatakis2019poincare,perolat2020poincar}. 
This raises our second question:
\begin{quote}
\emph{Can the pendulum-like cyclic behavior of FTRL dynamics in zero-sum games be understood and generalized via passivity?}
\end{quote}

\paragraph{Our Contributions.}
We provide affirmative answers to both questions raised above. 
We show that any finitely passive learning dynamic\footnote{
In control theory literature, a passive operator usually has a storage function with a finite lower bound.
In this paper, storage functions can have a finite lower bound or not, depending on the types of the operators.
We emphasize those operators which have a storage function with a finite lower bound as being \emph{finitely} passive.
For the definition of finitely passive learning dynamic, see Definition \ref{def:finitely-passive-learning-dynamic}.} guarantees constant regret (Theorem~\ref{thm:finite-passive-constant-regret}).
By using the notion of \emph{convex conjugate} from convex analysis,
we show that any continuous-time FTRL dynamic is finitely passive and lossless (Theorem~\ref{thm:cont-FTRL-finite-lossless});
the same holds for certain \emph{escort replicator dynamics}~\cite{Harper2011} (Appendix~\ref{app:escort}).
These generalize~\cite{Mabrok2018} which showed that Replicator Dynamic is finitely lossless.
Combining the two theorems above immediately recovers the result in \cite{GeorgiosSODA18} that any FTRL dynamic guarantees constant regret.
We note that in the analysis of Mabrok~\cite{Mabrok2018}, the state space (i.e.~the domain of the storage function) is the space of mixed strategies,
while we turn to a new state space of \emph{cumulative payoffs} for FTRL.
This choice is crucial for the generalization to FTRL dynamics, and it permits a cleaner proof via the tools established in convex analysis.

A key observation that enables the above results is that FTRL dynamic admits a storage function with a simple gradient structure,
which will be described formally in Section \ref{sect:convex}.
This motivates us to study a new family of lossless learning dynamics,
which is in one-one correspondence with the family of storage functions possessing that gradient structure.
By observing that such storage functions are closed under convex combination,
we discover that any convex combination of FTRL dynamics is finitely passive and lossless, and thus guarantees constant regret (Theorem \ref{thm:convex-combine-ftrl}).
``Convex combination of FTRL dynamics'' means:
Suppose there are $k$ FTRL dynamics, indexed from $1$ to $k$.
When we use the $j$-th one, it converts the cumulative payoffs at time $t$ to a mixed strategy $\bbx_j^t$.
A convex combination of these FTRL dynamics converts the cumulative payoffs at time $t$ to the mixed strategy $\sum_{j=1}^k \alpha_j \cdot \bbx_j^t$,
where $\alpha_j$'s are positive constants satisfying $\sum_{j=1}^k \alpha_j = 1$.

Convex combinations of lossless dynamics are directly analogous to linear combinations of conservative vector fields
in analyzing physical dynamics (see Figure \ref{fig:physics}).
This technique is also of practical relevance, since we might want to mix-and-match different dynamics to elicit their advantages.
For instance, different learning dynamics may lean toward either \emph{exploitation} or \emph{exploration}.
By combining them via convex combination with our own choice of $\alpha_j$'s, we can control our desired balance between exploitation and exploration (see Example \ref{ex:exploit-explore}).

We also show that for every graphical constant-sum game (e.g.~two-person zero-sum game) that admits a fully-mixed Nash equilibrium,
it corresponds to a finitely lossless game operator (Proposition~\ref{pr:game-oper-lossless}).
Thus, the game dynamic of any convex combinations of FTRL dynamics in such a game corresponds to a finitely lossless operator.
We use this observation to show that the game dynamic is almost perfectly recurrent, via the notion of \emph{Poincar\'{e} recurrence} (Theorem~\ref{thm:Poincare-convexFTRL}).
This distinguishes our work from \cite{fox2013population} and its subsequent works about learning in games,
as they mostly concern stability/convergence, while we study recurrence.

\paragraph{Roadmap.} In Sections~\ref{sect:prelim} and~\ref{sect:passivity}, we present the necessary background for this work, including the definitions of different operators,
the notions of (lossless) passivity and storage function, and some basic results about passivity.
In Section~\ref{sect:technical}, we show the desirable properties (finitely losslessness, constant regret) of FTRL dynamics.
In Section~\ref{sect:convex}, we present a characterization of lossless learning dynamics via the above-mentioned gradient structure of storage functions, and we discuss some properties of convex combinations of such learning dynamics.
The results about Poincar\'{e} recurrences of learning in graphical constant-sum games are presented in Section \ref{sect:recurrence}.
All missing proofs can be found in the appendix.
\section{Preliminary}\label{sect:prelim}

In this section, we define the operators depicted in Figure \ref{fig:FIC-simplified}.
We first define learning dynamic and its corresponding learning operator.
When there are multiple agents and each agent is using one learning dynamic,
their learning operators can be concatenated naturally to form a \emph{merged learning operator}.
Then we define game operator, whose interconnection with a merged learning operator in the manner of Figure \ref{fig:FIC-simplified}
is called a \emph{dynamical game system}.

We use a bold lower case to denote a vector variable.
Let $\rrplus := [0,+\infty)$ denote the set of non-negative real numbers.
In this paper, every function from $\rrplus$ to $\rr^d$ is assumed to be square integrable, and we call it a \emph{function-of-time}.
An (input-output) \emph{operator} is a mapping whose input and output are both functions-of-time.
Let $\Delta^n$ denote the probability simplex over $n$ actions,
i.e.~$\Delta^n := \{(x_1,\ldots,x_n)~\big|~ \sum_{j=1}^n x_j = 1;~\text{for}~1\le j\le n,~x_j\ge 0\}$.
$\inner{\bba}{\bbb}$ denotes the inner product of the vectors $\bba,\bbb$ of same dimension $d$, i.e.~$\inner{\bba}{\bbb} = \sum_{j=1}^d a_j b_j$.

\paragraph{Learning Dynamic and Learning Operator.}
We focus on the following type of continuous-time learning dynamics. An agent has an action set $A$; let $n := |A|$.
The process starts at time $0$. For any time $t\ge 0$ and for each action $j\in A$,
the agent computes the \emph{cumulative payoff} if she chooses action $j$ in the time interval $[0,t]$, denoted by $q_j(t)$.
Formally, for each action $j$, let $p_j(\tau)$ denote the \emph{(instantaneous) payoff} to action $j$ at time $\tau$. Then
\[
q_j(t) := q_j^0 + \int_0^t p_j(\tau)\,\mathsf{d}\tau,
\]
where $q_j^0$ is a constant chosen at the beginning of the process.
Let $\bbq(t) := (q_1(t),\cdots,q_n(t))$.
For any $t\ge 0$, the agent uses a \emph{conversion function} $f$ which takes $\bbq(t)$ as input,
and outputs a mixed strategy $\bbx(t)\in \Delta^n$ over the $n$ actions.
The process can be expressed compactly as an ordinary differential equations (ODE) system:
\begin{align}
\bbq(0) &= \bbq^0 &\text{\small (Initial condition)} \nonumber\\ 
\dot \bbq(t) &= \bbp(t) &\text{\small (Cumulative payoff/state update)} \label{eq:learning-algorithm-general} \\
\bbx(t) &= f(\bbq(t)). &\text{\small (Behavioral/strategy output)}\nonumber
\end{align}
When $\bbq^0 = (q_1^0,\ldots,q_n^0)$ is already given, the conversion function $f$ specifies the learning dynamic.
The learning dynamic can be viewed as an operator, which takes the function-of-time $\bbp : \rrplus \ra \rr^n$ as input,
and the output is another function-of-time $\bbx:\rrplus \ra \Delta^n$.

\paragraph{Regret.} For any $\bbp : \rrplus \ra \rr^n$, the \emph{regret} of a learning dynamic at time $T > 0$ is
\[
\left(\max_{j\in A}\int_0^T p_j(\tau)\,\mathsf{d}\tau\right)
-\int_0^T \inner{\bbp(\tau)}{\bbx(\tau)}\,\mathsf{d}\tau.
\]
We say a learning dynamic \emph{guarantees constant regret} if for any $\bbp : \rrplus \ra \rr^n$ and any $T>0$,
the regret at time $T$ is bounded from above by a constant that depends on $\bbq^0$ only.

\paragraph{Replicator Dynamic and FTRL Learning Dynamics.} When the conversion function $f$ is the logit choice map
\begin{equation}\label{eq:conversion-RD}
f_{\text{RD}}(\bbq) = \left( \frac{\exp(q_1)}{\calN} ~,~ \frac{\exp(q_2)}{\calN} ~,~ \cdots ~,~ \frac{\exp(q_n)}{\calN} \right),
\end{equation}
where $\calN = \sum_{j=1}^n \exp(q_j)$, the learning dynamic \eqref{eq:learning-algorithm-general} is equivalent to the well-known Replicator Dynamic \cite{HS98,Sandholm10},
which is the continuous-time analogue of Multiplicative Weights Update.

A FTRL learning dynamic is specified by a strictly convex \emph{regularizer function} $h:\Delta^{n}\ra \rr$, which determines the conversion function $f$ as below:
\begin{equation}\label{eq:conversion-FTRL}
f(\bbq) ~=~ \argmax_{\bbx\in \Delta^n} \left\{~\inner{\bbq}{\bbx} - h(\bbx)~\right\}.
\end{equation}
It is known that Replicator Dynamic is a special case of FTRL, by setting $h(\bbx) = \sum_{j=1}^{n} x_j \log x_j$.
Online Gradient Descent (OGD) is another commonly studied learning dynamic that is also a special case of FTRL with $L^2$ regularization,
i.e.~$h(\bbx) = \frac12 \sum_{j=1}^{n} (x_j)^2$.~\cite{hazan2016introduction}
When $n=2$, the conversion function of OGD is
\begin{equation}\label{eq:conversion-OGD}
f_{\text{OGD}}(\bbq) ~=~ \begin{cases}
(1,0) & \text{if } q_1 - q_2 \ge 1;\\
\left( \frac{q_1-q_2+1}{2} , \frac{q_2-q_1+1}{2} \right) & \text{if } 1 > q_1-q_2 > -1;\\
(0,1) & \text{if } q_1 - q_2 \le -1.
\end{cases}
\end{equation}

\paragraph{Merged Learning Operator.}
When a system has $m\ge 2$ agents, and each agent uses a learning dynamic,
we concatenate the corresponding learning operators together to form a \emph{merged learning operator} (MLO).
Precisely, the input to the MLO is $\hbbp = (\bbp^1,\cdots,\bbp^m)$, and its output is $\hbbx = (\bbx^1,\cdots,\bbx^m)$,
where $\bbx^i$ is the output of the learning operator of agent $i$ when its input is $\bbp^i$.
In this paper, we use hat notations (e.g.~$\hbbp,\hbbx$) to denote variables formed by such concatenations of variables of individual agents.

\paragraph{Game Operator and Dynamical Game System.}
Here, we provide a general definition of game operators (as in Figure \ref{fig:FIC-simplified}),
and leave the discussion about \emph{graphical constant-sum games}, which appear in our Poincar\'{e} recurrence results,
to Section \ref{sect:recurrence}.

A game has $m$ agents. Each agent $i$ has $n_i$ actions.
After each agent chooses a mixed strategy over her own actions, the game determines a payoff vector
$\hbbp\in \rr^{n_1}\times \cdots \times \rr^{n_m}$, where $\hat{p}_{k\ell}$ is the payoff to action $\ell$ of agent $k$.
Let $\Delta := \Delta^{n_1}\times \cdots \times \Delta^{n_m}$, and $\calP := \rr^{n_1}\times \cdots \times \rr^{n_m}$.
We can think of the game as a function $G:\Delta \ra \calP$.
Its game operator takes a function-of-time $\hbbx:\rrplus \ra \Delta$ as input,
and it outputs a function-of-time $\hbbp:\rrplus \ra \calP$, where $\hbbp(t) = G(\hbbx(t))$ for all $t\ge 0$.

A \emph{dynamical game system} (DGS) comprises of $m$ agents.
The game operator has input $\hbbx$ and output $\hbbp$.
Each agent uses a learning dynamic of the form \eqref{eq:learning-algorithm-general}.
The agents' learning operators are concatenated to form a MLO, which has input $\hbbp = (\bbp^1,\cdots,\bbp^m)$ and output $\hbbx = (\bbx^1,\cdots,\bbx^m)$
when $\hat{\bbr}\equiv \mathbf{0}$.
The MLO is interconnected with the game operator in the manner of Figure \ref{fig:FIC-simplified}.
\section{Passivity}\label{sect:passivity}

To motivate the notions of passivity and energy, consider an electrical network connected to a power source,
where the voltage and current across the network at time $\tau$ are respectively $v(\tau)$ and $i(\tau)$.
Let $E(t)$ denote the energy stored in the network at time $t$. We have
$E(t)\le E(0) + \int_0^t v(\tau) \cdot i(\tau) \,d\tau$.
The reason for the inequality (but not an exact equality) is that energy might dissipate from the network.
In this setting, the function-of-time $v$ is the input, while the function-of-time $i$ is the output,
so the network is indeed an operator; and as we shall see, this operator is passive.

\paragraph{Passivity of State Space System.}
To generalize the above idea to passivity of an ODE system, we need several mathematical notations.
Let $\bblt$ denote the Hilbert space of square integrable functions mapping $\rrplus$ to $\rr^n$
with inner product: $\inner{f}{g}_T := \int_0^T \inner{f(t)}{g(t)}\,\mathsf{d}t$.
Let $\bblte := \{ f:\rrplus \ra\rr^n~\big|~\inner{f}{f}_T < \infty~\text{for all }T\in \rrplus \}$.
An (input-output) operator is simply a mapping $S: \uu \ra \yy$, where $\uu,\yy\subset \bblte$.

We consider the following type of operators, which can be represented by
an ODE system called \emph{state space system} (SSS) of the following general form:
\begin{align}
\bbz(0) &~=~ \bbz^0 ;\nonumber\\
\dot \bbz(t) &~=~ g_1(\bbz(t),\bbu(t)) ;\label{eq:SSS} \\
\bby(t) &~=~ g_2(\bbz(t),\bbu(t)),\nonumber
\end{align}
where $\bbz^0\in \calZ\subset \rr^{d_1}$,
$\bbz:\rrplus\ra \calZ$, and $\bbu,\bby:\rrplus\ra\rr^{d_2}$. 
The set $\calZ$ is called the set of \emph{states}.
As the notations suggest, $\bbu,\bby$ are the input and output of this operator respectively.
When $\bbu$ is fed into the operator, the first two equalities define a well-posed ODE system,
so a unique solution of $\bbz$ exists under mild conditions.
Then $\bby$ is the output determined by a function $g_2$ of the unique solution $\bbz$ and the input $\bbu$.
The learning dynamic \eqref{eq:learning-algorithm-general} is such an operator, by viewing $\bbq,\bbp,\bbx,f(\bbq(t))$ in \eqref{eq:learning-algorithm-general}
as $\bbz,\bbu,\bby,g_2(\bbz(t),\bbu(t))$ in \eqref{eq:SSS} respectively.
We are ready to present the definition of passivity for such operators.

\begin{definition}\label{def:SSS}
A SSS is \emph{passive} if there exists a \emph{storage function} $L:\calZ\ra \rr$ such that
for all $\bbz^0\in \calZ$, $t\in \rrplus$ and all input-output pairs $\bbu\in \uu,~\bby\in \yy$, we have 
\begin{equation}\label{eq:passive-SSS}
L(\bbz(t)) ~\le~ L(\bbz^0) + \int_0^t \inner{\bbu(\tau)}{\bby(\tau)}\,\mathsf{d}\tau.
\end{equation}
If the equality always holds, then we say the SSS is \emph{lossless passive} or simply \emph{lossless}.
If a SSS is passive (resp.~lossless) via a storage function $L$ that has a finite lower bound,
we say it is \emph{finitely passive} (resp.~\emph{finitely lossless}).
\end{definition}

We note the reminiscence of inequality \eqref{eq:passive-SSS} with the inequality in the motivating example of electrical network.

When a SSS is finitely passive/lossless, we may assume, without loss of generality, that the finite lower bound of its storage function is zero.
We make this assumption on all finitely passive/lossless SSS in the rest of this paper.

\begin{figure}[t]
\centering \includegraphics[scale=0.95]{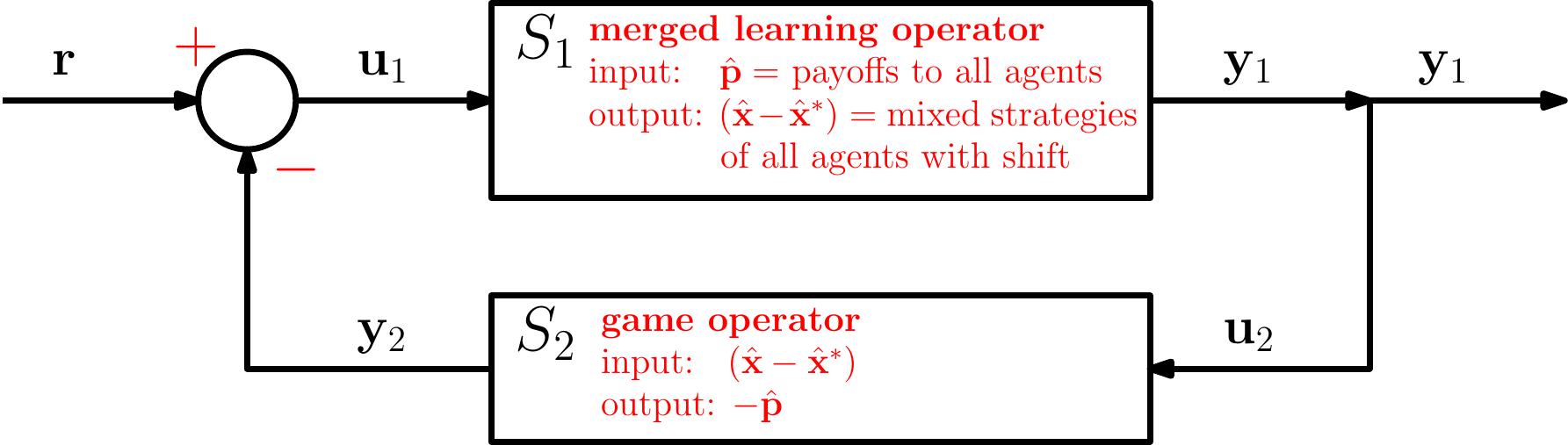}
\caption{A FIC system which includes two operators $S_1$ and $S_2$. We assume $\bbr \equiv \mathbf{0}$.
$S_1$ is a merged learning operator which converts payoffs to mixed strategies of agents with a shift $\hbbx^*$,
while $S_2$ is a game operator which converts the mixed strategies of agents (with shift $\hbbx^*$) to the negation of payoffs.
}\label{fig:FIC}
\end{figure}

\paragraph{Feedback Interconnection System.}
In Figure \ref{fig:FIC-simplified}, we presented a feedback interconnection (FIC) system.
While it is intuitive for readers to have a first understanding of the concept in control theory,
for our analysis it is more convenient to use Figure \ref{fig:FIC}.
The FIC system in Figure \ref{fig:FIC} consists of two SSS $S_1: \uu\ra \yy$ and $S_2: \yy\ra \uu$, with an external input source $\bbr\in \uu$,
while its output is $\bby_1\in \yy$; note that the FIC system is an operator by definition.
The variables are related via: $\bbu_1 = \bbr - \bby_2$, $\bby_1 = S_1(\bbu_1)$, $\bbu_2 = \bby_1$, and $\bby_2 = S_2(\bbu_2)$.

An important property of passive operators is that passivity is composable, i.e.~the composition of two passive operators results in a passive system.
Intuitively, if no operator in the total system is able to produce energy,
then the system as a whole cannot produce energy either.
The following theorem formally captures this intuition for FIC systems.

\begin{theorem}[\cite{fox2013population} Theorem 3.2]\label{thm:FIC-passive}
Consider the FIC system in Figure~\ref{fig:FIC}. Suppose that for $i=1,2$, $S_i$ is passive via storage function $L_i$.
Then the FIC system is a passive operator via storage function $L_1 + L_2$.
Precisely, for any $\bbz_1^0\in \calZ_1$, $\bbz_2^0\in \calZ_2$ and $t\in \rrplus$,
\[
L_1(\bbz_1(t)) + L_2(\bbz_2(t)) ~\le~ L_1(\bbz_1^0) + L_2(\bbz_2^0) + \int_0^t \inner{\bbr(\tau)}{\bby_1(\tau)}\,\mathsf{d}\tau.
\]
If $S_1,S_2$ are lossless, then the FIC system is lossless via storage function $L_1 + L_2$, i.e.~the inequality above becomes an equality.
\end{theorem}

Note that in the above theorem, if the FIC system is lossless and $\bbr$ is the zero function, then we have
$L_1(\bbz_1(t)) + L_2(\bbz_2(t)) = L_1(\bbz_1^0) + L_2(\bbz_2^0)$ for all $t\ge 0$,
i.e.~the value of $L_1+L_2$ does not change over time. The underlying dynamic is said to admit a \emph{constant-of-motion}.

\paragraph{DGS as a FIC System.} In the context of DGS, Figure \ref{fig:FIC} is obtained after several modifications from Figure \ref{fig:FIC-simplified}:
\begin{itemize}
\item In Figure \ref{fig:FIC-simplified}, the game operator's output is $\hbbp$, which is added to $\hat{\bbr}$ to form the MLO's input.
In Figure \ref{fig:FIC}, the game operator's output is $-\hbbp$ instead, and it is subtracted from $\bbr$ to form the MLO's input.
\item The MLO's output is $\hbbx$ in Figure \ref{fig:FIC-simplified}, but the MLO's output in Figure \ref{fig:FIC} has
a constant shift $\hbbx^* = (\bbx^{*,1},\ldots,\bbx^{*,m})\in \Delta$.
Precisely, the output is $\hbbx - \hbbx^*$. We call this operator the ``MLO with shift $\hbbx^*$''.
\item In Figure \ref{fig:FIC}, the game operator's input is $\hbbx -\hbbx^*$ instead of $\hbbx$,
while its output is $-\hbbp$ instead of $\hbbp$.
\end{itemize}

\paragraph{Basic Results about Passivity of Learning Operators.}
By viewing a DGS as a FIC system, we are interested in MLO and game operators which possess good properties like passivity.
The key advantage of this approach is it permits a DGS to be decoupled into two distinct operators, which can be analysed separately.

$S_1$ is a MLO. First, we show that it is passive if all learning operators possessed by the agents are passive.
This allows us to turn our focus to analyzing whether each individual learning operator is passive/lossless or not.

\begin{proposition}
Suppose for each agent $i$, her learning operator with shift $\bbx^{*,i}$ is passive (resp.~lossless) via a storage function $L^i$.
Then the MLO with shift $\hbbx^* = (\bbx^{*,1},\ldots,\bbx^{*,m})$ is passive (resp.~lossless) via the storage function $\sum_{i=1}^m L^i$.
\end{proposition}

By the next proposition, the choice of shift does \emph{not} affect passivity of the learning operator.
Thus, we say a learning dynamic is passive when its learning operator with \emph{any} shift is passive.

\begin{proposition}\label{prop:shift-no-effect}
Let $S^a,S^b$ be two learning operators of the same learning dynamic, with shifts $\bbx^{*,a},\bbx^{*,b}$ respectively.
$S^a$ is passive (resp.~lossless) via storage function $L^a(\bbq)$ if and only if
$S^b$ is passive (resp.~lossless) via storage function $L^a(\bbq) - \inner{(\bbx^{*,b}-\bbx^{*,a})}{\bbq} + c$, where $c$ is any constant.
\end{proposition}

The above proposition works even when the shifts are not mixed strategies, e.g.~when $\bbx^{*,a}$ is the zero vector.
We use $E$ to denote a storage function of a learning operator with zero shift,
and $L$ to denote a storage function of a learning operator with a shift of a mixed strategy.

While a shift does not affect passivity, it \emph{does} affect whether the learning operator is \emph{finitely} passive or not.
In order to prove that certain learning dynamics guarantee constant regret, we need their learning operators with some specific shifts to be finitely passive.
This motivates the following definition of \emph{finitely passive learning dynamics}.
Let $\bbe_j$ denote the vector with the $j$-th entry be one, and all other entries be zero.

\begin{definition}\label{def:finitely-passive-learning-dynamic}
A learning dynamic is finitely passive (resp.~finitely lossless) if for every action $j$,
its learning operator with shift $\bbe_j$ is finitely passive (resp.~finitely lossless).
\end{definition}

\begin{proposition}\label{prop:shift-mixed-strategy}
If a learning dynamic is finitely passive (resp.~finitely lossless), then for any mixed strategy $\bbx^*$,
the learning operator with shift $\bbx^*$ is finitely passive (resp.~finitely lossless).
\end{proposition}
\section{Passivity of Learning Operators}\label{sect:technical}

\subsection{FTRL Dynamics are Finitely Lossless}\label{sect:FTRL-lossless}

We start the analysis by establishing a strong connection between FTRL dynamics and passivity. Specifically, FTRL dynamics are finitely lossless.

\begin{theorem}\label{thm:cont-FTRL-finite-lossless}
Given any FTRL dynamic over $n$ actions and with regularizer function $h$, let the \emph{convex conjugate} of $h$ be
\begin{equation}\label{eq:h-star}
h^*(\bbq) := \max_{\bbx\in \Delta^n} \left\{ \inner{\bbq}{\bbx} - h(\bbx) \right\}.
\end{equation}
Then for any $\bbx^*\in \Delta^n$, the learning operator with shift $\bbx^*$ is finitely lossless via the storage function $L(\bbq)$ given below:
\begin{equation}\label{eq:storage-function-ftrl}
L(\bbq) ~=~ 
h^*(\bbq)- \inner{\bbq}{\bbx^*} + h(\bbx^*).
\end{equation}
In particular, for any action $j$, the learning operator with shift $\bbe_j$ is finitely lossless,
and hence any FTRL dynamic is finitely lossless by Definition \ref{def:finitely-passive-learning-dynamic}.
\end{theorem}

\begin{pf}
By the ``maximizing argument'' identity in p.~149 of \cite{Shalev2012}, we have $\nabla h^*(\bbq(t)) = \bbx(t)$.
Hence, $\nabla L(\bbq(t)) = \nabla h^*(\bbq(t)) - \bbx^* = \bbx(t) - \bbx^*$.
By the chain rule, $\difft{L(\bbq(t))} = \inner{\bbx(t) - \bbx^*}{\bbp(t)}$, and hence $L(\bbq(t)) = L(\bbq(0)) + \int_0^t \inner{\bbx(\tau) - \bbx^*}{\bbp(\tau)}\,\mathsf{d}\tau$, 
verifying that the operator is lossless.
Moreover, $L$ is bounded below by zero, since by the definition of $h^*$, for any $\bbq$,
$L(\bbq) = h^*(\bbq) - \inner{\bbq}{\bbx^*} + h(\bbx^*) \ge \left(\inner{\bbq}{\bbx^*} - h(\bbx^*)\right) - \inner{\bbq}{\bbx^*} + h(\bbx^*) = 0$.
\end{pf}

We summarize the properties of the operators of FTRL dynamics with various shifts in Figure~\ref{fig:comparisons}.

\begin{figure}[h]
\centering
\begin{tabular}{|c|c|c|}
\hline
&&\\ [-2.4ex]
& \bold{FTRL with zero shift}  & \bold{FTRL with shift of a mixed strategy $\bbx^*$} \\
\hline
&&\\ [-2.4ex]
\bold{input} & $\bbp$ (payoff)  & $\bbp$\\
\hline
&&\\ [-2.4ex]
\bold{output} & $\bbx$ (mixed strategy)  & $\bbx-\bbx^*$\\
\hline
&&\\ [-2.4ex]
\bold{state} & $\bbq$ (cumulative payoff) & $\bbq$ \\
\hline 
&&\\ [-2.4ex]
\bold{storage function} & $E(\bbq) = h^*(\bbq)$ (see \eqref{eq:h-star}) & $L(\bbq) = h^*(\bbq) - \inner{\bbq}{\bbx^*}+ h(\bbx^*)$ \\
\hline
&&\\ [-2.4ex]
\bold{infimum} & $-\infty$ & $0$ \\
\hline
&&\\ [-2.4ex]
\bold{property} & lossless & finitely lossless \\
\hline
\end{tabular}
\caption{FTRL learning operators with various shifts.}\label{fig:comparisons}
\end{figure}

\subsection{Relationship to Regret}

\begin{theorem}\label{thm:finite-passive-constant-regret}
Any finitely passive learning dynamic guarantees constant regret.
\end{theorem}

\begin{pf}
Let $L^j$ denote the storage function of the learning operator with shift $\bbe_j$. 
Since the learning dynamic is finitely passive, we can assume the infimum of $L^j$ is zero.
By the definition of passivity,
\[
L^j(\bbq(t)) ~\le~ L^j(\bbq(0)) + \int_0^t \inner{\bbx(\tau)-\bbe_j}{\bbp(\tau)}\,\mathsf{d}\tau.
\]
Hence, the regret w.r.t.~action $j$ at time $t$ satisfies:
\[
\int_0^t \inner{\bbe_j}{\bbp(\tau)}\,\mathsf{d}\tau ~-~ \int_0^t \inner{\bbx(\tau)}{\bbp(\tau)}\,\mathsf{d}\tau 
~\le~  L^j(\bbq(0)) - L^j(\bbq(t)) ~\le~ L^j(\bbq(0)).
\]
Thus, the regret up to time $t$ is bounded by $\max_j \left\{ L^j(\bbq(0)) \right\}$, which is a constant that depends only on the initial state $\bbq^0$.
\end{pf}

As an immediate corollary of Theorems \ref{thm:cont-FTRL-finite-lossless} and \ref{thm:finite-passive-constant-regret},
all FTRL dynamics guarantee bounded regret.

\begin{corollary}[\cite{GeorgiosSODA18}]
Every FTRL dynamic guarantees constant regret.
\end{corollary}

Theorem \ref{thm:finite-passive-constant-regret} states that finite passivity implies constant regret.
The following proposition states that the converse (constant regret implies finite passivity) is also true, if we restrict to lossless learning dynamics.

\begin{proposition}\label{pr:finitely-lossless-necessary}
Suppose that a learning dynamic is lossless. Then the learning dynamic guarantees constant regret if and only if it is finitely lossless.
\end{proposition}

\section{A Characterization of Lossless Learning Dynamics, and Their\\Convex Combinations}\label{sect:convex}

In Section~\ref{sect:FTRL-lossless}, we showed that every FTRL dynamic is lossless. A FTRL dynamic is specified by a convex regularizer function $h$,
while the storage function $L$ has an \emph{implicit} form in term of $h$.
Here, we ask how we can specify the storage function directly to \emph{generate} a lossless learning dynamic.
Not all storage functions work, so we first seek some necessary conditions on them. 
These conditions are stated using the storage function $E$ of the learning operator with zero shift.
By Proposition~\ref{prop:shift-no-effect}, the lossless storage function of the learning operator with shift $\bbx^*$ is
$E(\bbq)-\inner{\bbq}{\bbx^*} + c$.

Suppose there is a lossless learning dynamic in the form of \eqref{eq:learning-algorithm-general};
recall that when $\bbq^0$ is already given, the learning dynamic is specified by its conversion function $f$.
Let $E$ be the storage function of its learning operator with zero shift. We present two necessary conditions on $E$ and $f$.

\medskip

\parabold{Necessary Condition 1 (NC1).} (i) For any $j=1,\cdots,n$, $\nabla_j E(\bbq) \ge 0$. (ii) For any $\bbq\in \rr^n$, we have $\nabla E(\bbq) = f(\bbq)$.

\noindent\underline{Reason:} 
Since the learning dynamic is lossless, $E(\bbq(t)) = E(\bbq^0) + \int_0^t \inner{\bbx(\tau)}{\bbp(\tau)}\,\mathsf{d}\tau$.
Taking time-derivative on both sides yields
\[
\difft{E(\bbq(t))} = \inner{\bbx(t)}{\bbp(t)} = \inner{f(\bbq(t))}{\bbp(t)}.
\]
On the other hand, by the chain rule, we have
\[
\difft{E(\bbq(t))} = \inner{\nabla E(\bbq(t))}{\dot \bbq} = \inner{\nabla E(\bbq(t))}{\bbp(t)}.
\]
Thus, $\inner{f(\bbq(t))}{\bbp(t)} = \inner{\nabla E(\bbq(t))}{\bbp(t)}$ for any $\bbp(t)$ and $\bbq(t)$.
This readily implies conditions (ii) holds.\footnote{To formally argue this,
for any $\bbq\in \rr^n$ and for each action $j$, we construct continuous $\bbp: [0,t]\ra \rr^n$
such that $\bbq^0 + \int_0^t \bbp(\tau)\,\mathsf{d}\tau = \bbq$ and $\bbp(t) = \bbe_j$.
This implies that the $j$-th component of $f(\bbq)$ is same as the $j$-th component of $\nabla E(\bbq)$.
The construction is easy to make.}
Since $f(\bbq)\in \Delta^n$, condition (i) holds.

\medskip

\parabold{Necessary Condition 2 (NC2).} For any real number $r$ and any $\bbq$, $E(\bbq+r\cdot \bbone) = E(\bbq) + r$.

\noindent\underline{Reason:} By NC1, $\nabla E(\bbq) = f(\bbq)$, which is in $\Delta^n$.
Thus, the directional derivative of $E$ along the direction $\bbone$ is $\inner{\nabla E(\bbq)}{\bbone} = \inner{f(\bbq)}{\bbone} =1$.

\medskip

Indeed, it is easy to verify that the above two necessary conditions are also sufficient.

\begin{proposition}\label{prop:characterize-lossless}
Any smooth function $E(\bbq)$ which satisfies NC1(i) and NC2
is a lossless storage function of a learning operator with zero shift.
The conversion function $f$ satisfies $f(\bbq) = \nabla E(\bbq)$.
\end{proposition}

\begin{pf}
Due to NC2, $\inner{\nabla E(\bbq)}{\bbone} = 1$, thus $\sum_{j=1}^n \nabla_j E(\bbq) = 1$.
This equality and NC1(i) implies $\nabla E(\bbq)\in \Delta^n$.
Now, consider a learning dynamic with conversion function $f = \nabla E$. Then for any function-of-time $\bbp$ and any $t>0$,
we have $\difft{E(\bbq(t))} = \inner{\nabla E(\bbq(t))}{\dot \bbq} = \inner{f(\bbq(t))}{\bbp(t)} = \inner{\bbx(t)}{\bbp(t)}$.
Integrating both sides w.r.t.~$t$ shows that the learning dynamic is lossless via the storage function $E$.
\end{pf}

\begin{example}
Consider the learning dynamic where the conversion function always outputs the uniform mixed strategy,
i.e.~for all $\bbq$, $f(\bbq) = \frac 1n \cdot \mathbf{1}$.
By Proposition \ref{prop:characterize-lossless}, it is a lossless learning dynamic via $E(\bbq) = \inner{\frac 1n \cdot \mathbf{1}}{\bbq}$,
a storage function that clearly satisfies NC1(i) and NC2.
Of course, there is not really ``learning'' with this conversion function, and it is easy to construct examples to demonstrate that its regret can be unbounded.
By Proposition \ref{pr:finitely-lossless-necessary}, this learning dynamic is not finitely lossless.
This shows the family of finitely lossless learning dynamics is a \emph{proper} subset of the family of lossless learning dynamics.
\end{example}

The family of smooth functions $E$ satisfying NC1(i) and NC2 can be represented compactly as
\[
\calE := \{E:\rr^n\ra \rr~|~\forall \bbq,~\nabla E(\bbq) \ge \mathbf{0}~~\text{and}~~\inner{\nabla E(\bbq)}{\bbone} = 1 \}.
\]
An interesting observation is that this family is closed under \emph{convex combination},
i.e.~if $E_1,\ldots,E_k\in \calE$, then for any real numbers $\alpha_1,\ldots,\alpha_k \ge 0$ such that $\sum_{\ell=1}^k \alpha_\ell = 1$,
$\left(\sum_{\ell=1}^k \alpha_\ell \cdot E_\ell\right)\in \calE$.
By Proposition \ref{prop:characterize-lossless}, $E_1,\ldots,E_k$ are lossless storage functions of some learning operators with zero shift.
Suppose the conversion functions are $f_1,\ldots,f_k$ respectively.
Then by Proposition \ref{prop:characterize-lossless} again, $\left(\sum_{\ell=1}^k \alpha_\ell \cdot E_\ell\right)$
is a lossless storage function of a learning operator with zero shift, with conversion function $\left(\sum_{\ell=1}^k \alpha_\ell \cdot f_\ell\right)$.
This motivates the following definition of \emph{convex combination of learning dynamics}.

\begin{definition}\label{def:convex-combination}
Given $k$ learning dynamics, each over $n$ actions, let $f_\ell$ denote the conversion function of the $\ell$-th learning dynamic.
Given any non-negative constants $\alpha_1,\cdots,\alpha_k$ where $\sum_{\ell=1}^k \alpha_\ell = 1$,
the convex combination of the $k$ learning dynamics with parameters $\alpha_1,\cdots,\alpha_k$
is a learning dynamic with conversion function $\left(\sum_{\ell=1}^k \alpha_\ell \cdot f_\ell\right)$.
\end{definition}

\begin{example}\label{ex:exploit-explore}
Suppose we are using the half-half convex combination of Replicator Dynamic (RD) and Online Gradient Descent (OGD), and $n=2$. 
Recall the conversion functions of RD and OGD in \eqref{eq:conversion-RD} and \eqref{eq:conversion-OGD}.
Suppose that at some time $t$, $\bbq(t) = (0.6,0)$. Then the mixed strategy at time $t$ with the half-half convex combination of RD and OGD is
\begin{align*}
\bbx(t) &~=~ \frac 12 \cdot f_{\text{\emph{RD}}} \left((0.6,0)\right) + \frac 12 \cdot f_{\text{\emph{OGD}}} \left((0.6,0)\right)\\
&~=~ \frac 12 \left[ \left(\frac{\exp(0.6)}{\exp(0.6) + \exp(0)} ~,~ \frac{\exp(0)}{\exp(0.6) + \exp(0)}\right) + \left( \frac{0.6-0+1}{2}~,~ \frac{0-0.6+1}{2} \right) \right]\\
&~\approx~ \frac 12 \left[ (0.6457,0.3543) + (0.8,0.2) \right] ~\approx~ (0.7228,0.2772).
\end{align*}

By \eqref{eq:conversion-OGD}, $f_{\text{\emph{OGD}}}(\bbq)$ outputs a strategy with zero probability of choosing action 2 whenever $q_1-q_2\ge 1$.
In contrast, $f_{\text{\emph{RD}}}(\bbq)$ maintains a tiny but positive probability of choosing action 2 even when $q_1$ is much larger than $q_2$.
We may say OGD leans toward \emph{exploitation} while RD leans toward \emph{exploration}.
By combining the two learning dynamics via convex combination with our own choice of $\alpha$'s,
we obtain a new lossless learning dynamic with our desired balance between exploitation and exploration.
\end{example}

Convex combination not only preserves losslessness, but also finitely losslessness.
Suppose there are several finitely lossless learning dynamics. By Definition \ref{def:finitely-passive-learning-dynamic},
for every action $j$, the storage functions of their learning operators with shift $\bbe_j$ have finite lower bounds.
It is easy to verify that for a convex combination of these lossless learning dynamics, its learning operator with shift $\bbe_j$
is lossless via the same convex combination of the storage functions mentioned above.
Since the storage functions have finite lower bounds, their convex combination has a finite lower bound too.

\begin{theorem}\label{thm:convex-combination}
Given any $k$ lossless (resp.~finitely lossless) learning dynamics,
any convex combination of them is a lossless (resp.~finitely lossless) learning dynamic.
\end{theorem}

Theorems~\ref{thm:cont-FTRL-finite-lossless},~\ref{thm:finite-passive-constant-regret} and~\ref{thm:convex-combination} lead to the interesting theorem below.

\begin{theorem}\label{thm:convex-combine-ftrl}
Any convex combination of any finitely lossless learning dynamics is a learning dynamic that guarantees constant regret.
In particular, any convex combination of any FTRL learning dynamics is a learning dynamic that guarantees constant regret.
\end{theorem}

\noindent\textbf{Remark.} 
Note that the family of smooth storage functions that generate \emph{finitely} lossless learning dynamics is
\[
\left\{ E\in \calE~|~\forall j,~E(\bbq)-\inner{\bbq}{\bbe_j}~\text{is bounded from below}\right\}.
\]
Again, this family is closed under convex combination.
By Theorem \ref{thm:cont-FTRL-finite-lossless}, the family of FTRL learning dynamics is a subset
of the family of finitely lossless learning dynamics.
It is not clear if the two families are equal; we believe not.
For instance, for the half-half convex combination of Replicator Dynamic and Online Gradient Descent,
we cannot find any regularizer function that validates it is a FTRL dynamic.
\section{Lossless DGS and Poincar\'{e} Recurrence}\label{sect:recurrence}

In the last section, we present a potentially broad family of (finitely) lossless learning dynamics.
In this section, our focus is on DGS in which agents use such learning dynamics.
We first prove that for certain \emph{graphical constant-sum games}, their game operators are finitely lossless.
Thus, for any DGS comprising of such game and finitely lossless learning dynamics,
it admits a constant-of-motion by Theorem \ref{thm:FIC-passive} when $\bbr \equiv 0$,
i.e.~the sum of the two storage function values is a constant for any $t\ge 0$.
Then we use this constant-of-motion and follow a principled approach proposed by Mertikopoulos et al.~\cite{GeorgiosSODA18} to show our main result here:
the DGS is Poincar\'{e} recurrent. 
In the rest of this section, we first define graphical constant-sum game and Poincar\'{e} recurrence,
then we apply the principled approach to prove our main result.

\paragraph{Graphical Constant-sum Game.}
There are $m$ agents. Each agent $i$ has $n_i$ actions.
In a \emph{graphical game} \cite{Kearns01}, there is a normal-form (matrix) game between every pair of agents, which we call an \emph{edge-game}.
The edge game between agents $i,k$ is specified by two matrices, $\bbA^{ik}\in \rr^{n_i\times n_k}$ and $\bbA^{ki}\in \rr^{n_k\times n_i}$.
When each agent $k\neq i$ chooses a mixed strategy $\bbx_k \in \Delta^{n_k}$,
the payoff vector of agent $i$, which contains the payoff to each action of agent $i$,
is the sum of the payoff vectors in all her edge-games. Precisely, the payoff vector of agent $i$ is
\begin{equation}\label{eq:graphical-game}
\bbp_i ~=~ \sum_{1\le k\le m,~k\neq i}~
\bbA^{ik} \cdot \bbx_k.
\end{equation}
A \emph{graphical constant-sum game} \cite{DP09} is a graphical game such that
for every pair of agents $\{i,k\}$, there exists a constant $c^{\{i,k\}}$ satisfying
$\bbA^{ik}_{j\ell} + \bbA^{ki}_{\ell j} = c^{\{i,k\}}$ for any action $j$ of agent $i$ and action $\ell$ of agent $k$.

\paragraph{Game Operator.}
Recall that a game operator has an input of mixed strategies of different agents with shift $\hbbx^*$,
while the output is the negative of the cumulative payoff vector to different actions.
We point out one useful fact, which follows quite readily from~\cite{piliouras2014optimization}.

\begin{proposition}\label{pr:game-oper-lossless}
If $\hbbx^*$ is a Nash equilibrium of a graphical constant-sum game, then the game operator with shift $\hbbx^*$ is passive; the storage function is the zero function.
Moreover, if $\hbbx^*$ is fully-mixed (i.e.~every entry in $\hbbx^*$ is positive), then the game operator is lossless via the zero storage function.
\end{proposition}

When the game is lossless, by the composability of passive operators (Theorem~\ref{thm:FIC-passive}),
if it is coupled with passive (resp.~lossless) learning dynamics, then the DGS is passive (resp.~lossless).

\paragraph{Poincar\'{e} Recurrence.}
In a DGS, $\hbbp$ is a function of $\hbbx$ via the game operator, while $\hbbx$ is a function of $\hbbq$ via the conversion function.
Thus, $\hbbp$ is a function of $\hbbq$. We say the ODE system~\eqref{eq:learning-algorithm-general} is \emph{divergence-free} if
$\sum_{i=1}^m \sum_j \frac{\partial p_{ij}}{\partial q_{ij}}$ is zero everywhere.
When $\hbbp$ is derived using a graphical game via \eqref{eq:graphical-game},
$p_{ij}$ does not depend on $q_{ij}$ since for every $k\neq i$, $\bbx_k$ is a function of $\bbq_k$ only. Thus, the game dynamic is divergence-free.

Intuitively, an ODE system with domain $\rr^N$ is Poincar\'{e} recurrent if almost all trajectories return arbitrarily close to their initial position infinitely often. 
In order to work formally with the notion of Poincar\'{e} recurrence we need to define a measure on $\rr^N$.
We use the standard Lebesgue measure on $\rr^N$.
Liouville's formula states that divergence-free ODE systems preserve volume~\cite{Weibull};
see Appendix~\ref{app:Liouville} for more discussion of volume and Liouville's formula.
Thus, the following Poincar\'{e} Recurrence Theorem is applicable if its bounded-orbit requirement is satisfied.

\begin{theorem}[\cite{Poincare1890,barreira}]\label{thm:Poincare}
If a transformation preserves volume and has bounded orbits, then it is Poincar\'{e} recurrent,
i.e.~for each open set there exist orbits that intersect this set infinitely often.
\end{theorem}

Given any $\ep>0$, we can cover $\rr^n$ by countably many balls of radius $\ep$, and apply the theorem to each ball.
We conclude that almost every point returns to within an $\epsilon$ neighbourhood of itself. Since $\epsilon>0$ is arbitrary,
we conclude that almost every initial point is almost recurrent.

\paragraph{Poincar\'{e} Recurrence in DGS.}
Recall that in each learning operator of a DGS, the state is represented by a vector $\bbq \in \rr^n$,
which represents the cumulative payoffs. Clearly, it can be unbounded as time goes even in a graphical constant sum game\footnote{For instance,
this happens for a two-person zero-sum game with every payoff entry to agent 1 is strictly positive.}, thus prohibiting us to use Theorem \ref{thm:Poincare}.
Instead, as in \cite{GeorgiosSODA18}, we consider a transformation that maps $\bbq = (q_1,\ldots,q_n)$ to 
\[
(q_1',q_2',\cdots,q_{n-1}') ~:= (q_1 - q_n,\ldots,q_{n-1}-q_n)\in \rr^{n-1}.
\]
It is well-known that for any FTRL dynamic with starting point $\bbq^0 = (q^0_1,\ldots,q^0_n)$ and conversion function $f:\rr^n \ra \Delta^n$,
it is equivalent to the following dynamic with state variables $\bbq' = (q_1',\ldots,q_{n-1}')\in \rr^{n-1}$,
whereas $1\le j \le n-1$:
\begin{align}
q'_j(0) &~=~ q^0_j - q^0_n \nonumber\\
\dot q'_j(t) &~=~ p_j(t)-p_n(t) \nonumber\\
\bbx(t) &~=~ f(q'_1(t),q'_2(t),\cdots,q'_{n-1}(t),0).\label{eq:conversion-reduced-dual}
\end{align}
Given a mixed strategy $\bbx^*$, if we cast the output of the above dynamic to be $(x_1(t)-x_1^*,\cdots,x_{n-1}(t)-x_{n-1}^*)$,
then it is easy to verify that the learning operator is finitely lossless
via the storage function $\overline{L}(q'_1,\ldots,q'_{n-1}) = L(q'_1,\cdots,q'_{n-1},0)$,
where $L$ is the storage function of the original learning operator with shift $\bbx^*$.

\begin{theorem}\label{thm:Poincare-convexFTRL}
Poincar\'{e} recurrence occurs in the strategy space $\Delta$ for any dynamical game system where
(1) each agent employs a learning dynamic which is a convex combination of FTRL; and
(2) the underlying game is a graphical constant-sum game with a fully-mixed Nash equilibrium.
\end{theorem}

\begin{figure}[t]
\centering\includegraphics[scale=0.3]{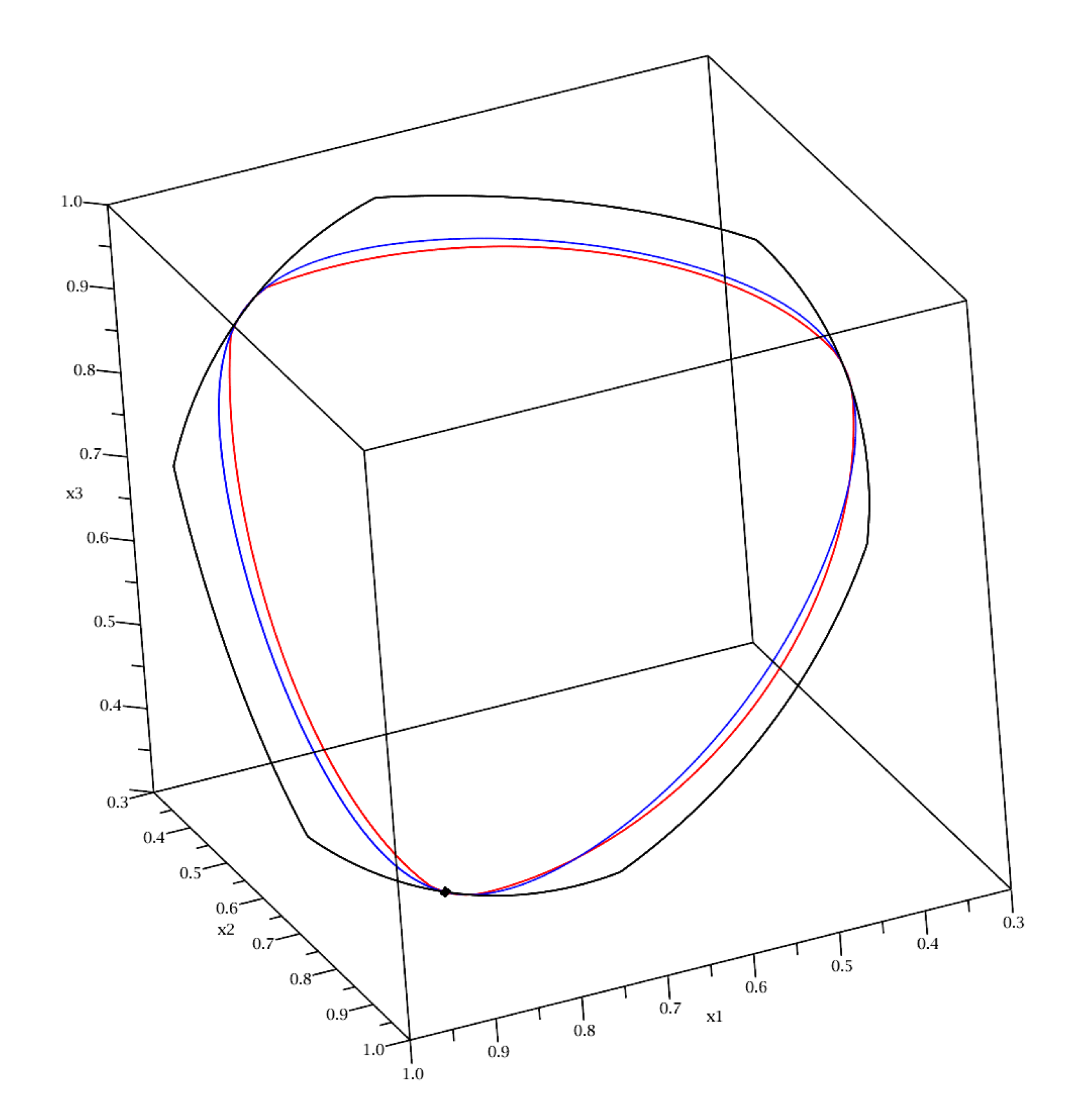}
\caption{The trajectories of learning in a graphical constant-sum game called \emph{Cyclic Matching Pennies}.
In this game, there are three agents, each has two actions. $\bbA^{12} = \bbA^{23} = \bbA^{31} = 
\begin{bmatrix}
1 & 0\\
0 & 1
\end{bmatrix}$, while $\bbA^{21} = \bbA^{32} = \bbA^{13} = 
\begin{bmatrix}
-1 & 0\\
0 & -1
\end{bmatrix}$. The mixed strategies of the three agents are denoted by $(x_1,1-x_1)$, $(x_2,1-x_2)$ and $(x_3,1-x_3)$ respectively.
All trajectories start with $x_1=0.9$, $x_2=0.88$ and $x_3=0.4$ (the black dot in the figure).
The trajectories are simulated when all agents use Replicator Dynamics (blue), Online Gradient Descent (Black),
and half-half convex combination of the former two dynamics (red) respectively. All trajectories are recurrent.}\label{fig:poincare-3d}
\end{figure}

See Figure~\ref{fig:poincare-3d} for an illumination of Poincar\'{e} recurrence under conditions (1) and (2).
To prove the theorem, we first show that Poincar\'{e} recurrence occurs in the space that contains $\hat{bbq}'$
($\hat{\bbq}'$ is the concatenation of the $\bbq'$ of all agents).
This comprises of two major steps:
\begin{itemize}
\item The dynamic preserves \emph{volume}, since the dynamic is divergence-free.
\item For any starting point $\hat{\bbq}'(0)$, the dynamic remains bounded.
To show this, we use the fact that $\overline{L}$ is a constant-of-motion of the game dynamic,
so for any $t\ge 0$, $\hat{\bbq}'(t)$ must stay within a level set of $\overline{L}$, which is bounded (see Appendix~\ref{app:recurrence}).
\end{itemize}
Poincar\'{e} recurrence in the space that contains $\hat{\bbq}'$ implies Poincar\'{e} recurrence in the strategy space $\Delta$,
since the conversion function in~\eqref{eq:conversion-reduced-dual} is continuous.

\section{Conclusion}

We present a control-theoretic perspective to understanding popular learning dynamics like FTRL and escort replicator dynamics.
At the heart of it is the use of storage (energy) functions to govern how the dynamic turns history of payoffs to strategic choices.
This mirrors the study of physical dynamics, e.g.~electrical networks \cite{khalil2015nonlinear}.
Analysis via storage functions permits us to prove optimal regret bounds and inspires interesting generalizations of FTRL via convex combinations.

An important benefit of these control-theoretic tools is, as pointed out by Fox and Shamma~\cite{fox2013population},
they allow decoupling of game dynamics into learning operators and game operators. 
This provides a framework to understand learning-in-games via a modular approach, by analyzing these operators separately.
This technique can liberate us from analyzing each individual learning-in-game system in ad-hoc manner.

In our work, we initiate the study of connections between online optimization and control theory with continuous-time learning dynamics.
An interesting problem is how such a connection can be generalized to discrete-time learning \emph{algorithms},
e.g. Multiplicative Weights Update and its optimistic variant~\cite{BP2018,Cheung2018,DP2019,CP2020}.
There does exist theory that generalizes passivity to discrete-time settings, e.g.~Section VI.7 in \cite{Desoer1975}.
We hope our work inspires further studies in this direction.

Lastly, we believe that this control-theoretic perspective is also useful for understanding learning dynamics/algorithms which are not always passive.
The perspective can help us spot under which situations the learning dynamics/algorithms create or dissipate energy.
By avoiding situations where energy is created, it is possible that we can achieve stable outcomes in learning processes.

\section*{Acknowledgements}

This research/project is supported in part by NRF2019-NRF-ANR095 ALIAS grant, grant PIE-SGP-AI-2018-01, NRF 2018 Fellowship NRF-NRFF2018-07,
AME Programmatic Fund (Grant No.~A20H6b0151) from the Agency for Science, Technology and Research (A*STAR) and the National Research Foundation,
Singapore under its AI Singapore Program (AISG Award No: AISG2-RP-2020-016).

\bibliographystyle{alpha}
\bibliography{sigproc2,IEEEabrv,Bibliography,refer}

\newpage

\appendix

\section*{\LARGE Appendix}

\section{Volume and Liouville's Formula}\label{app:Liouville}

Let $g:\rr^d \ra \rr^d$ be a function. Given an ODE system $\dot \bbz = g(\bbz)$ but with a flexibility to choose the starting point,
let $\Phi(\bbz^0,t)$ be the solution of the ODE system at time $t$ with starting point $\bbz^0$.
Given any set $A$, let $A(t) = \{\Phi(\bbz^0,t) ~|~ \bbz^0 \in A\}$. When $A$ is measurable, under mild conditions on the ODE system,
$A(t)$ is measurable and its volume is $\text{vol}[A(t)]= \int_{A(t)}\,\mathsf{d}v$.
Liouville's formula states that the time derivative of the volume $A(t)$ is equal to the integral of the divergence of the ODE system over $A(t)$:
\[
\frac{d}{dt} \text{vol}[A(t)] = \int_{A(t)} \text{trace}\left( \frac{\partial g}{\partial \bbz} \right) \,\mathsf{d}v,
\]
where $\frac{\partial g}{\partial \bbz}$ is the \emph{Jacobian} of the ODE system.
Note that $\text{trace}\left( \frac{\partial g}{\partial \bbz} \right) = \sum_{j=1}^d \frac{\partial g_j}{\partial z_j}$,
where $g_j$ is the $j$-th component of the function $g$.
This immediately implies volume preservation for divergence-free systems.

\section{Missing Proofs}\label{app:missing}

\begin{pfof}{Proposition~\ref{prop:shift-no-effect}}
It suffices to prove the forward (only if) direction, as the other direction is symmetric. By the definition of passivity,
\[
L^a(\bbq(t)) \le L^a(\bbq^0) + \int_0^t \inner{(\bbx(\tau)-\bbx^{*,a})}{\bbp(\tau)}\,\mathsf{d}\tau.
\]
This implies
\begin{align*}
L^a(\bbq(t)) &~\le~ L^a(\bbq^0) + \int_0^t \inner{(\bbx(\tau)-\bbx^{*,b})}{\bbp(\tau)}\,\mathsf{d}\tau + \int_0^t \inner{(\bbx^{*,b}-\bbx^{*,a})}{\bbp(\tau)}\,\mathsf{d}\tau\\
&~=~  L^a(\bbq^0) + \int_0^t \inner{(\bbx(\tau)-\bbx^{*,b})}{\bbp(\tau)}\,\mathsf{d}\tau + \inner{(\bbx^{*,b}-\bbx^{*,a})}{(\bbq(t)-\bbq^0)}.
\end{align*}
Thus, by setting $L^b(\bbq) := L^a(\bbq) - \inner{(\bbx^{*,b}-\bbx^{*,a})}{\bbq} + c$, we have 
\[
L^b(\bbq(t)) \le L^b(\bbq^0) + \int_0^t \inner{(\bbx(\tau)-\bbx^{*,b})}{\bbp(\tau)}\,\mathsf{d}\tau,
\]
certifying passivity of the operator $S^b$.
\end{pfof}

\begin{pfof}{Proposition~\ref{prop:shift-mixed-strategy}}
Suppose that for each action $j$, the learning operator with shift $\bbe_j$ is finitely passive via storage function $L^j$.
We claim that the storage function $\sum_{j=1}^n x^*_j \cdot L^j$ can certify finitely passivity of
the learning operator with shift of the mixed strategy $\bbx^*$.
To see why, note that for each action $j$, we have
\[
L^j(\bbq(t)) \le L^j(\bbq^0) + \int_0^t \inner{(\bbx(\tau)-\bbe_j)}{\bbp(\tau)}\,\mathsf{d}\tau.
\]
We are done by multiply both sides of the inequality by $x^*_j$, then summing up over all $j$'s.
\end{pfof}

\newpage

\begin{pfof}{Proposition~\ref{pr:finitely-lossless-necessary}}
$(\Leftarrow)$ Done by Theorem~\ref{thm:finite-passive-constant-regret}.

\noindent $(\Rightarrow)$ Suppose the contrary, i.e.~the learning algorithm is lossless,
but there exists $j$ such that the learning operator with shift $\bbe_j$ is \emph{not} finitely lossless.
Thus, it has a storage function $L^j$ which is not bounded from below, and
\[
L^j(\bbq(t)) ~=~ L^j(\bbq(0)) + \int_0^t \inner{\bbx(\tau)}{\bbp(\tau)}\,\mathsf{d}\tau - \int_0^t \inner{\bbe_j}{\bbp(\tau)}\,\mathsf{d}\tau.
\]
Following the calculation in the proof of Theorem~\ref{thm:finite-passive-constant-regret}, the regret w.r.t.~action $j$ at time $t$
is \emph{exactly} equal to $L^j(\bbq^0) - L^j(\bbq(t))$.

Since $L^j$ is not bounded from below, for any $r < 0$, there exists $\tilde{\bbq}$ such that $L^j(\tilde{\bbq}) \le r$.
It is easy to construct $\bbp$ such that $\bbq(t) = \bbq^0 + \int_0^t \bbp(\tau)\,\mathsf{d}\tau = \tilde{\bbq}$;
for instance, set $\bbp(\tau) = (\tilde{\bbq}-\bbq^0)/t$ for all $\tau\in [0,t]$.
For this choice of $\bbp$, the regret at time $t$ is $L^j(\bbq^0) - L^j(\tilde{\bbq}) \ge L^j(\bbq^0) - r$.
Since we can choose arbitrarily negative value of $r$, the learning dynamic cannot guarantee constant regret, a contradiction.
\end{pfof}

\begin{pfof}{Proposition~\ref{pr:game-oper-lossless}}
To show that the game operator is passive, according to Definition~\ref{def:SSS} and the input-output choice of $S_2$ (see Figure~\ref{fig:FIC}),
it suffices to show that
\[
\int_0^t \inner{(\hbbx(\tau)-\hbbx^*)}{(-\hbbp(\tau))}\,\mathsf{d}\tau
~=~
-\int_0^t \underbrace{\inner{\hbbx(\tau)}{\hbbp(\tau)}}_{V_1}\,\mathsf{d}\tau + \int_0^t \underbrace{\inner{\hbbx^*}{\hbbp(\tau)}}_{V_2}\,\mathsf{d}\tau ~\ge~ 0.
\]
Recall the definition of $c^{\{i,k\}}$ in a graphical constant-sum game.
Since $V_1$ is simply the total payoffs to all agents, $V_1$ is the sum of the constants $c^{\{i,k\}}$ of all edge-games,
i.e.~$V_1 = \sum_{i=1}^{m-1} \sum_{k=i+1}^m c^{\{i,k\}}$. We denote this double summation by $V$.
It remains to show that $V_2 \ge V$ always if we want to show the game operator is passive,
and to show that $V_2 = V$ always if we want to show the game operator is lossless.

Let the action set of agent $i$ be $S_i$. We first expand $V_2 = \inner{\hbbx^*}{\hbbp}$ as follows:
\[
\inner{\hbbx^*}{\hbbp} ~=~ \sum_{i=1}^m \sum_{j\in S_i} x_{ij}^* \sum_{\substack{k=1\\k\neq i}}^m [\bbA^{ik} \bbx_k]_j 
~=~ \sum_{i=1}^m \sum_{\substack{k=1\\k\neq i}}^m (\bbx_i^*)\trans \bbA^{ik} \bbx_k
~=~ \sum_{i=1}^m \sum_{\substack{k=1\\k\neq i}}^m (\bbx_k)\trans (\bbA^{ik})\trans \bbx_i^*~;
\]
the last equality holds since $(\bbx_i^*)\trans \bbA^{ik} \bbx_k$ is the transpose of $(\bbx_k)\trans (\bbA^{ik})\trans \bbx_i^*$.
Then we rewrite the double summation on the RHS as follows:
\begin{align*}
\sum_{i=1}^m \sum_{\substack{k=1\\k\neq i}}^m (\bbx_k)\trans (\bbA^{ik})\trans \bbx_i^*
&~=~ \sum_{i=1}^m \sum_{\substack{k=1\\k\neq i}}^m \left[ c^{\{i,k\}} - (\bbx_k)\trans \bbA^{ki} \bbx_i^* \right] \comm{definition of constant-sum edge-game}\\
&~=~ \sum_{i=1}^m \sum_{\substack{k=1\\k\neq i}}^m c^{\{i,k\}}
~-~ \sum_{k=1}^m \sum_{\substack{i=1\\i\neq k}}^m (\bbx_k)\trans \bbA^{ki} \bbx_i^* \\
&~=~ 2V ~-~ \sum_{k=1}^m \underbrace{\sum_{\substack{i=1\\i\neq k}}^m (\bbx_k)\trans \bbA^{ki} \bbx_i^*}_{U_k}.
\end{align*}

It remains to bound the term $\sum_{k=1}^m U_k$. Observe that for each agent $k$,
$U_k$ is the payoff to agent $k$ when she chooses the mixed strategy $\bbx_k$, while every other agent $i$ chooses the mixed strategy $\bbx_i^*$.
Since the mixed strategies $\bbx_i^*$ are coming from a Nash equilibrium (NE),
by the definition of NE, $U_k \le v_k^*$, where $v_k^*$ is the payoff to agent $k$ at the NE.
Thus, $\sum_{k=1}^m U_k \le \sum_{k=1}^m v_k^*$, where the RHS is the total payoffs to all agents at the NE.
Since the game is constant-sum, we have $\sum_{k=1}^m v_k^* = V$. Hence, $V_2 = \inner{\hbbx^*}{\hbbp} \ge 2V - V = V$.

When the NE is fully-mixed, we have the following extra property: 
at the NE, for the agent $k$, her payoff from each of her actions is the same, and equals to $v_k^*$.
Thus, $U_k$ exactly equals to $v_k^*$, so $V_2 = V$.
\end{pfof}
\section{Poincar\'{e} Recurrence}\label{app:recurrence}

We first formally state the following corollary of Proposition~\ref{pr:game-oper-lossless} and Theorem~\ref{thm:FIC-passive}.

\begin{corollary}\label{co:lossless-FIC}
	The FIC system which corresponds to a dynamical game system, in which $S_1$ is any finitely lossless MLO
	and $S_2$ is any game operator which corresponds to a graphical constant-sum game with a fully-mixed Nash equilibrium,
	is finitely lossless. The storage function that demonstrates finitely losslessness of the FIC system is the same as the storage function of $S_1$.
	When the external input $\bbr$ is the zero function, the storage function becomes a constant-of-motion.
\end{corollary}

To complete the proof of Theorem \ref{thm:Poincare-convexFTRL}, we need to show the second property required by the principled approach of \cite{GeorgiosSODA18}.
It relies crucially on the following lemma. Recall that we have defined the following in the main paper,
which converts the storage function for the original learning operator to the storage function of the new learning operator of \eqref{eq:conversion-reduced-dual}.
\begin{equation}\label{eq:Lbar}
\overline{L}(q'_1,q'_2,\cdots,q'_{n-1}) = L(q'_1,q'_2,\cdots,q'_{n-1},0),
\end{equation}

\begin{lemma}[Adapted from \cite{GeorgiosSODA18}, Appendix D]\label{lem:bounded-level-set}
	For any continuous FTRL dynamic and for any $\bbx^*\in \Delta^n$, 
	let $L$ be its finitely lossless storage function defined in~\eqref{eq:storage-function-ftrl},
	and let $\overline{L}$ be the function defined on $\rr^{n-1}$ as in~\eqref{eq:Lbar}.
	Then any level set of $\overline{L}$ is bounded in $\rr^{n-1}$,
	i.e.~for any real number $\bar{c}$, the set below is bounded:
	\[
	\{~(q'_1,\cdots,q'_{n-1}) ~\big|~\overline{L}(q'_1,\cdots,q'_{n-1}) \le \bar{c}~\}.
	\]
\end{lemma}

Recall the definition of FTRL and Theorem~\ref{thm:cont-FTRL-finite-lossless}.
For each agent $i$, suppose she uses a convex combination of $\ell_i$ FTRL dynamics indexed by $i1,i2,\cdots,i\ell_i$.
Let the storage functions of these FTRL dynamics be $L^{i1},L^{i2},\cdots,L^{i\ell_i}$.
Also, let $\bbq'^{,i}$ denote a vector in $\rr^{n_i-1}$ for agent $i$.
Then the storage function of the whole dynamical game system is
\[
\sum_{i=1}^m \sum_{j=1}^{\ell_i} \alpha_{ij} \cdot \overline{L}^{ij} (\bbq'^{,i}),~~~\text{where}~\alpha_{ij} > 0,~\text{and}~\forall i,~\sum_{j=1}^{\ell_i} \alpha_{ij} = 1.
\]
Due to Corollary~\ref{co:lossless-FIC}, this storage function is a constant-of-motion when $\bbr\equiv 0$,
and thus is bounded by certain constant $\bar{c}$ when the starting point is already given.
Since every $L^{ij}$ and hence $\overline{L}^{ij}$ has infimum zero, we must have: for each agent $i$, $\alpha_{i1}\cdot \overline{L}^{i1}(\bbq'^{,i}) \le \bar{c}$,
and hence $\overline{L}^{i1}(\bbq'^{,i}) \le \bar{c}/\alpha_{i1}$.
Then by Lemma~\ref{lem:bounded-level-set}, for each agent $i$, $\bbq'^{,i}(t)$ remains bounded for all $t$,
and thus the overall vector $\hbbq'(t) = (\bbq'^{,1}(t),\bbq'^{,2}(t),\cdots,\bbq'^{,m}(t))$ also remains bounded for all $t$.
\section{Escort Learning Dynamics}\label{app:escort}

An escort learning dynamic~\cite{Harper2011} is a system of differential equations on variable $\bbx\in \Delta^n$: for each $1\le j\le n$,
\[
\dot x_j ~=~ \phi_j(x_j) \cdot \left[ p_j - \frac{\sum_{\ell=1}^n \phi_\ell(x_\ell) \cdot p_\ell}{\sum_{\ell=1}^n \phi_\ell(x_\ell)} \right],
\]
where each $\phi_j$ is a positive function on domain $(0,1)$. Note that when $\phi_j(x_j) = x_j$, this is Replicator Dynamic.

\begin{proposition}
Suppose a learning dynamic has the following property: if it starts at a point in the interior of $\Delta^n$, then it stays in the interior forever.
We have:
the learning dynamic is FTRL via a separable strictly convex regularizer function $h(\bbx) = \sum_{i=1}^n h_i(\bbx_i)$ if and only if it is an escort replicator dynamic.
\end{proposition}

\begin{pf}
\newcommand{\bx}{\bar{x}}
If the specified learning dynamic is FTRL, recall that the conversion function takes $\bbq$ as input, and output the mixed strategy
\[
\argmax_{\bbx\in \Delta^{n}} \left\{ \inner{\bbq}{\bbx} - h(\bbx)  \right\},
\]
which we denote by $\bbx(\bbq)$ in this proof.
Let $\bx_j = 1/h_j''(x_j)$ and $H := \sum_j \bx_j$. When $\bbx(\bbq)$ is in the interior of $\Delta^n$ for some $\bbq$,
by Appendix D of \cite{cheung2019vortices}, we have
\[
\frac{\partial x_j}{\partial q_j} ~=~ \bx_j - \frac{[\bx_j]^2}{H}
~~~~~~~~\text{and}~~~~~~~~
\forall \ell\neq j,~~
\frac{\partial x_j}{\partial q_\ell} ~=~ -\frac{\bx_j \bx_\ell}{H}.
\]
By the chain rule,
\[
\dot x_j ~=~ \left[ \bx_j - \frac{[\bx_j]^2}{H} \right] \cdot p_j ~+~ \sum_{\ell\neq j} \left[ -\frac{\bx_j \bx_\ell}{H} \right] p_\ell
~=~ \bx_j \left( p_j - \frac{\sum_{\ell=1}^n \bx_\ell p_\ell}{\sum_{\ell=1}^n \bx_\ell} \right).
\]
By recognizing $\phi_j(x_j)$ as $\bx_j$, the FTRL dynamic is an escort replicator dynamic.
Precisely, we set $\phi_j(x_j) = 1/h_j''(x_j)$. Since $h$ is strictly convex, $h_j''$ is a positive function, hence $\phi_j$ is a positive function too.

Conversely, if the specified algorithm is an escort learning dynamic with escort function $\phi_j$ for each $j$,
to show that it is a FTRL dynamic with some strictly convex regularizer $h$,
we set $h$ to be separable, and for each $j$, $h_j''(x_j) = 1/\phi_j(x_j)$.
Thus, it suffice to set $h_j$ to be any double anti-derivative of $1/\phi_j$.
Since $h_j''(x_j) = 1/\phi_j(x_j) > 0$, each $h_j$ is strictly convex, and hence $h$ is strictly convex.
\end{pf}

\section{More Plots Illuminating Poincar\'{e} Recurrences}\label{app:figures}

Finally, we present more plots that illuminate Poincar\'{e} recurrences of learning in games in the next two pages.

\begin{figure}[t]
\centering
\includegraphics[scale=0.67]{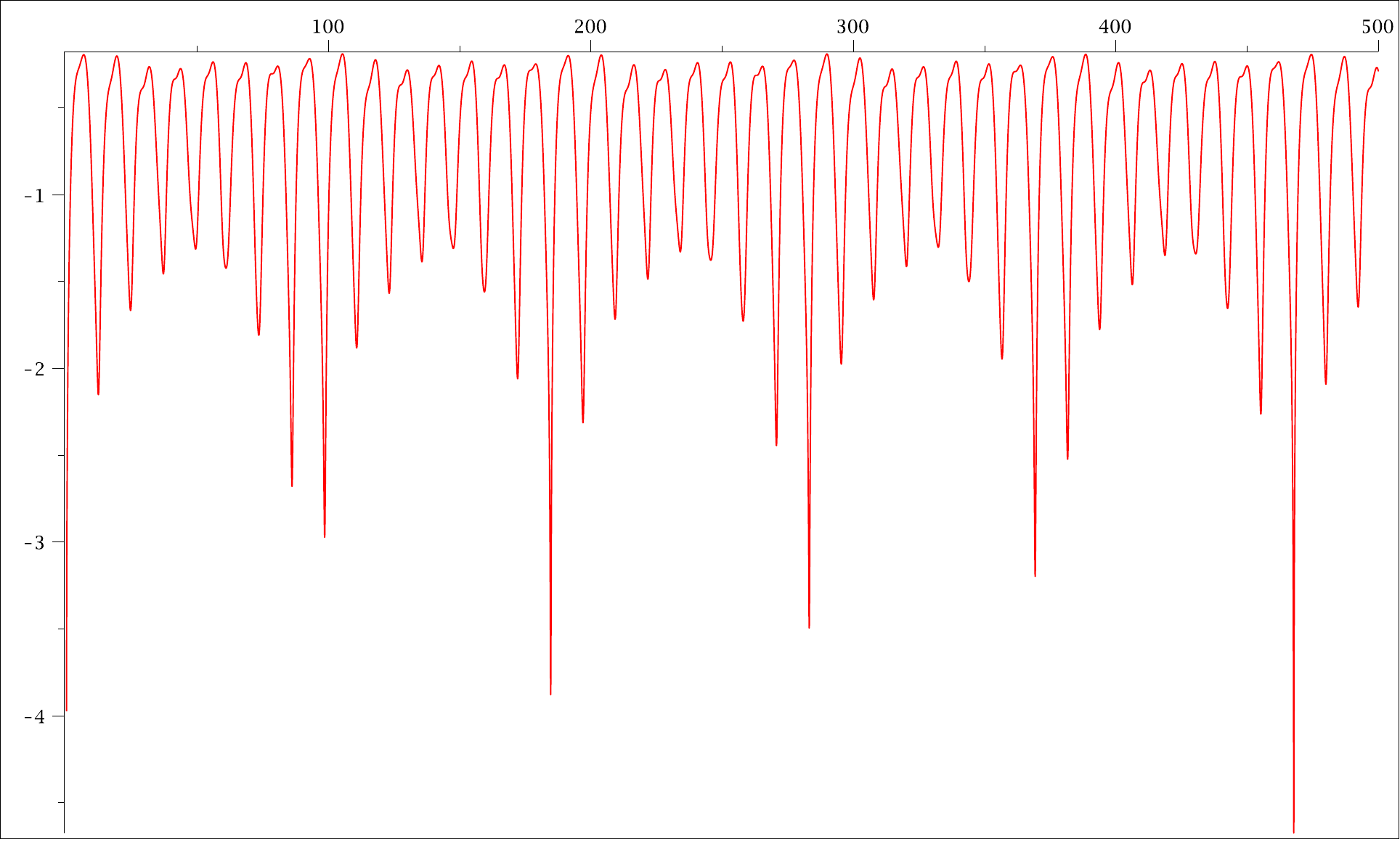}
\includegraphics[scale=0.67]{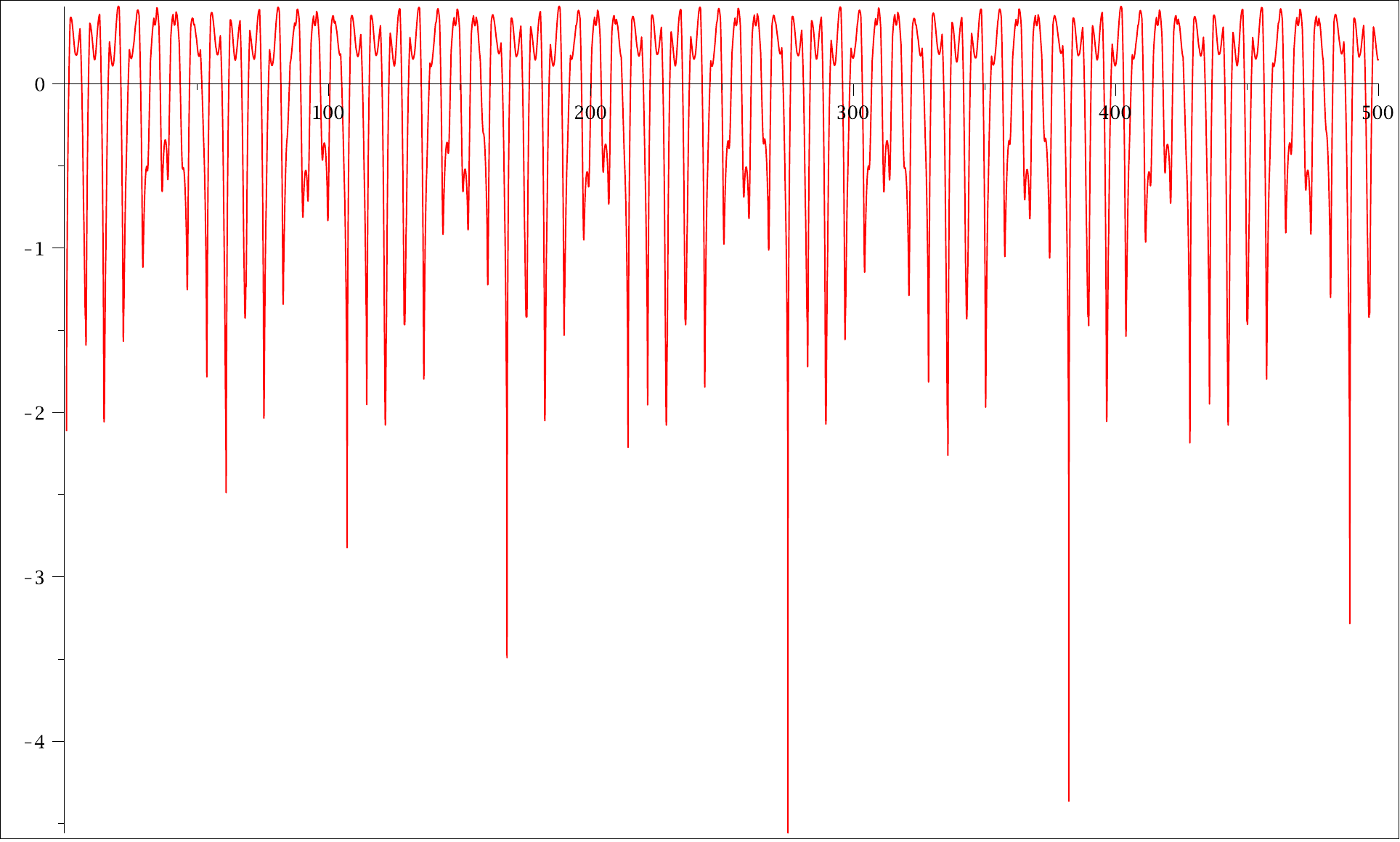}
\caption{Poincar\'{e} recurrences of \textsf{Replicator Dynamics} (\textsf{RD}; top) and \textsf{Online Gradient Descent} (\textsf{OGD}; bottom)
in the classical two-agent Rock-Paper-Scissors game.
agent 1 starts with mixed strategy $\bbx_1(0) = (0.5,0.25,0.25)$, while agent 2 starts with mixed strategy $\bbx_2(0) = (0.6,0.3,0.1)$.
The two graphs plot the logarithm of the Euclidean distance between $(\bbx^1(t),\bbx^2(t))$ and $(\bbx^1(0),\bbx^2(0))$, from $t=0.1$ to $t=500$.
Every downward spike corresponds to a moment where the flow gets back close to the starting point.
In both cases, the distance drops below $10^{-3}$ for multiple times.}\label{fig:poincare1}
\end{figure}

\begin{figure}[t]
\centering
\includegraphics[scale=0.7]{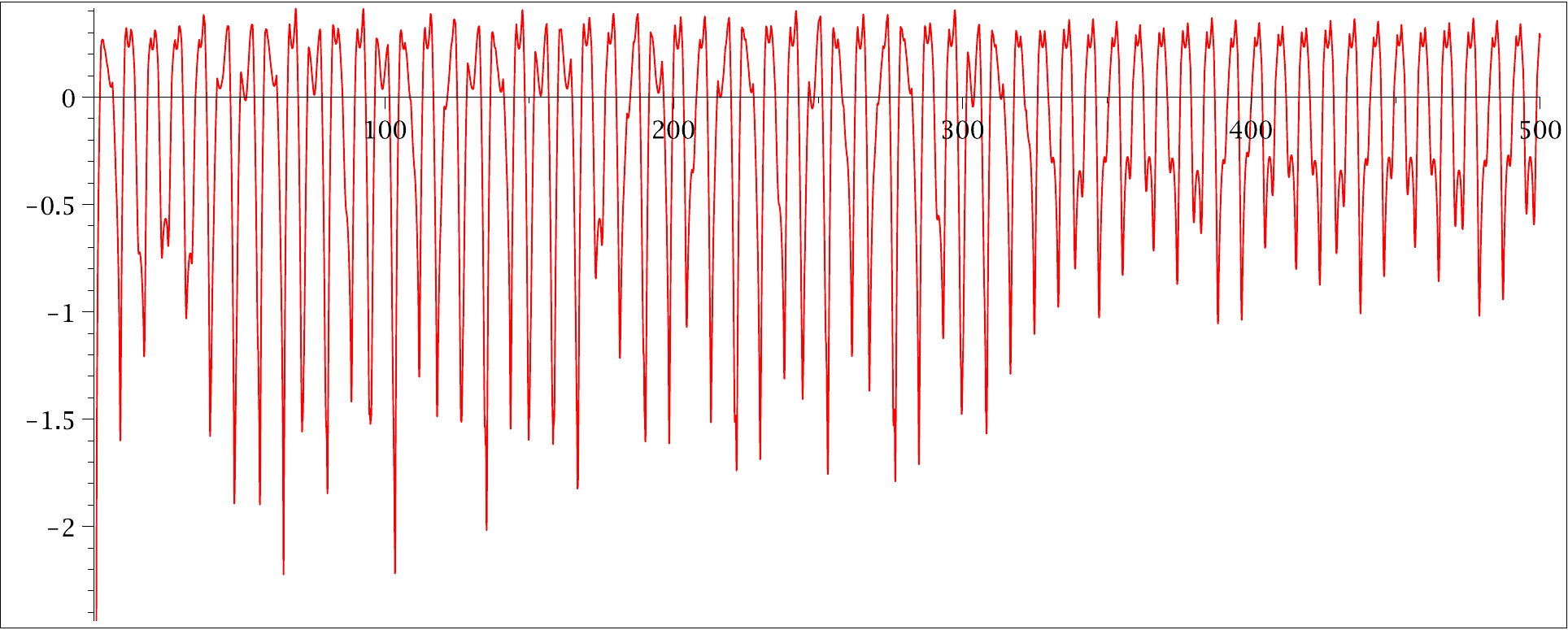}
\includegraphics[scale=0.7]{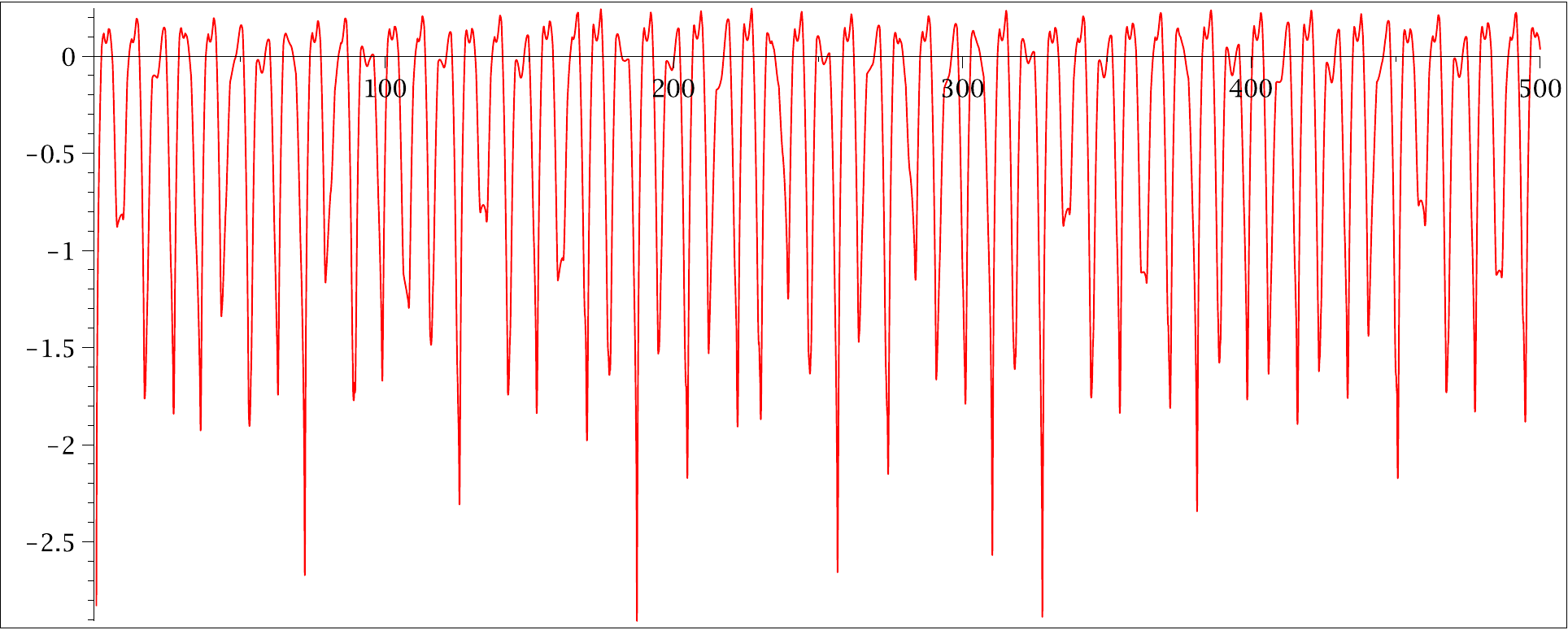}
\includegraphics[scale=0.7]{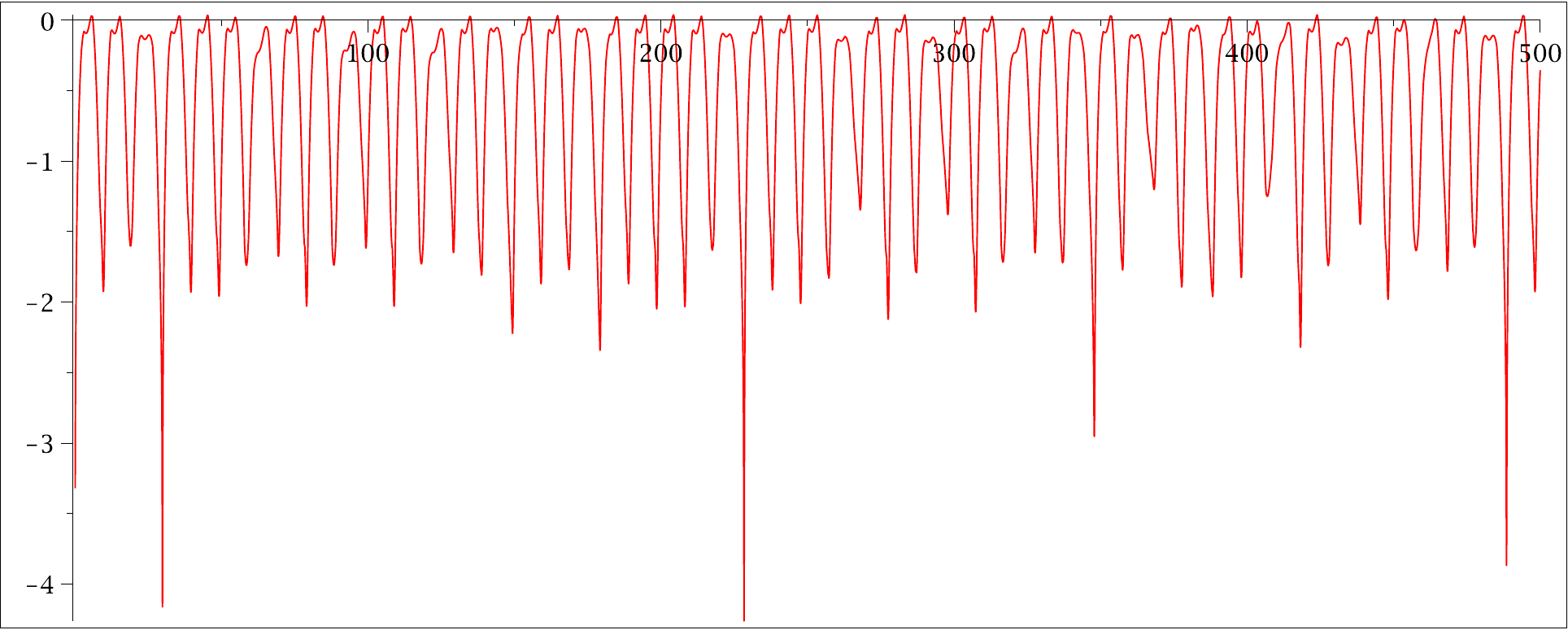}
\caption{Poincar\'{e} recurrence of $\alpha \cdot \textsf{RD} + (1-\alpha) \cdot \textsf{OGD}$ in the classical Rock-Paper-Scissors game,
for $\alpha = 1/4$ (top), $\alpha=1/2$ (middle) and $\alpha = 3/4$ (bottom).
The starting point is $(\bbx_1(0),\bbx_2(0)) = ((0.5,0.25,0.25),(0.6,0.3,0.1))$.
The graphs plot the logarithm of the Euclidean distance between $(\bbx_1(t),\bbx_2(t))$ and $(\bbx_1(0),\bbx_2(0))$, from $t=0.1$ to $t=500$.}\label{fig:poincare2}
\end{figure}

\end{document}